\documentclass[10pt,twocolumn,letterpaper]{article}

\usepackage{iccv}
\usepackage{times}
\usepackage{epsfig}
\usepackage{graphicx}
\usepackage{amsmath}
\usepackage{amssymb}

\usepackage{times}
\usepackage{epsfig}
\usepackage{graphicx}
\usepackage{amsmath}
\usepackage{amssymb}
\usepackage{siunitx}
\usepackage{multirow}
\usepackage{mycommands}
\usepackage{color}
\usepackage{pifont}
\newcommand{\cmark}{\ding{51}}
\newcommand{\xmark}{\ding{55}}
\usepackage{tabularx}
\usepackage{xpatch}
\usepackage{floatrow}

\usepackage{arydshln}
\usepackage{float}

\newcommand\blfootnote[1]{%
  \begingroup
  \renewcommand\thefootnote{}\footnote{#1}%
  \addtocounter{footnote}{-1}%
  \endgroup
}
\usepackage[hang,flushmargin]{footmisc}

\usepackage[pagebackref=true,breaklinks=true,letterpaper=true,colorlinks,bookmarks=false]{hyperref}

\iccvfinalcopy

\ificcvfinal\fi

\begin{document}
%%% TEMPLATE TRICKS
\newcommand{\mypartight}[1]{\noindent {\bf #1}}
\newcommand{\myparagraph}[1]{\vspace{3pt}\noindent\textbf{#1}\xspace}

%%% COLOR-CODING COMMENTS
\newcommand{\alert}[1]{{\color{red}{#1}}}
\newcommand{\blue}[1]{{\color{blue}{#1}}}
\newcommand{\red}[1]{{\color{red}{#1}}}
\newcommand{\gt}[1]{{\color{purple}{GT: #1}}}
\newcommand{\gtt}[1]{{\color{purple}{#1}}}
\newcommand{\gtr}[2]{{\color{purple}\st{#1} {#2}}}

\newcommand\boldblue[1]{\textcolor{blue}{\textbf{#1}}}

%  revisited oxford/paris notation
\def\roxf{$\mathcal{R}$Oxford\xspace}
\def\rox{$\mathcal{R}$Oxf\xspace}
\def\ro{$\mathcal{R}$O\xspace}
\def\rpar{$\mathcal{R}$Paris\xspace}
\def\rpa{$\mathcal{R}$Par\xspace}
\def\rp{$\mathcal{R}$P\xspace}
\def\rdis{$\mathcal{R}$1M\xspace}

\newcommand\resnet[3]{\ensuremath{\prescript{#2}{}{\mathtt{R}}{#1}_{\scriptscriptstyle #3}}\xspace}

%--------------------------------------------------------------------------------
% Method macros 
\newcommand{\ours}{Beat-them-all\xspace} % our method's acronym

%--------------------------------------------------------------------------------
% stdev in tables
\newcommand{\stddev}[1]{\scriptsize{$\pm#1$}}

\newcommand{\diffup}[1]{{\color{OliveGreen}{($\uparrow$ #1)}}}
\newcommand{\diffdown}[1]{{\color{BrickRed}{($\downarrow$ #1)}}}
%--------------------------------------------------------------------------------

%--------------------------------------------------------------------------------
%algorithm
\newcommand{\comment} [1]{} %{\color{orange} \Comment     #1}} % colored comment

%--------------------------------------------------------------------
% spaces: useful for tabulars
\def\nmsp{\hspace{-6pt}}
\def\nssp{\hspace{-3pt}}
\def\nxssp{\hspace{-1pt}}
\def\zsp{\hspace{0pt}}
\def\xssp{\hspace{1pt}}
\def\ssp{\hspace{3pt}}
\def\msp{\hspace{6pt}}
\def\lsp{\hspace{12pt}}
\def\xlsp{\hspace{20pt}}

%--------------------------------------------------------------------

\newcommand{\head}[1]{{\smallskip\noindent\bf #1}}
\newcommand{\equ}[1]{(\ref{equ:#1})\xspace}

%--------------------------------------------------------------------

\newcommand{\nn}[1]{\ensuremath{\text{NN}_{#1}}\xspace}
\def\l1{\ensuremath{\ell_1}\xspace}
\def\l2{\ensuremath{\ell_2}\xspace}

%--------------------------------------------------------------------

\newcommand{\tran}{^\top}
\newcommand{\mtran}{^{-\top}}
\newcommand{\zcol}{\mathbf{0}}
\newcommand{\zrow}{\zcol\tran}

\newcommand{\ind}{\mathds{1}}
\newcommand{\expect}{\mathbb{E}}
\newcommand{\nat}{\mathbb{N}}
\newcommand{\zahl}{\mathbb{Z}}
\newcommand{\real}{\mathbb{R}}
\newcommand{\proj}{\mathbb{P}}
\newcommand{\prob}{\mathbf{Pr}}

\newcommand{\mif}{\textrm{if }}
\newcommand{\other}{\textrm{otherwise}}
\newcommand{\minimize}{\textrm{minimize }}
\newcommand{\maximize}{\textrm{maximize }}

\newcommand{\id}{\operatorname{id}}
\newcommand{\const}{\operatorname{const}}
\newcommand{\sgn}{\operatorname{sgn}}
\newcommand{\var}{\operatorname{Var}}
\newcommand{\mean}{\operatorname{mean}}
\newcommand{\trace}{\operatorname{tr}}
\newcommand{\diag}{\operatorname{diag}}
\newcommand{\vect}{\operatorname{vec}}
\newcommand{\cov}{\operatorname{cov}}

\newcommand{\softmax}{\operatorname{softmax}}
\newcommand{\clip}{\operatorname{clip}}

\newcommand{\defn}{\mathrel{:=}}
\newcommand{\peq}{\mathrel{+\!=}}
\newcommand{\meq}{\mathrel{-\!=}}

\newcommand{\floor}[1]{\left\lfloor{#1}\right\rfloor}
\newcommand{\ceil}[1]{\left\lceil{#1}\right\rceil}
\newcommand{\inner}[1]{\left\langle{#1}\right\rangle}
\newcommand{\norm}[1]{\left\|{#1}\right\|}
\newcommand{\frob}[1]{\norm{#1}_F}
\newcommand{\card}[1]{\left|{#1}\right|\xspace}
\newcommand{\diff}{\mathrm{d}}
\newcommand{\der}[3][]{\frac{d^{#1}#2}{d#3^{#1}}}
\newcommand{\pder}[3][]{\frac{\partial^{#1}{#2}}{\partial{#3^{#1}}}}
\newcommand{\ipder}[3][]{\partial^{#1}{#2}/\partial{#3^{#1}}}
\newcommand{\dder}[3]{\frac{\partial^2{#1}}{\partial{#2}\partial{#3}}}

\newcommand{\wb}[1]{\overline{#1}}
\newcommand{\wt}[1]{\widetilde{#1}}

\newcommand{\cA}{\mathcal{A}}
\newcommand{\cB}{\mathcal{B}}
\newcommand{\cC}{\mathcal{C}}
\newcommand{\cD}{\mathcal{D}}
\newcommand{\cE}{\mathcal{E}}
\newcommand{\cF}{\mathcal{F}}
\newcommand{\cG}{\mathcal{G}}
\newcommand{\cH}{\mathcal{H}}
\newcommand{\cI}{\mathcal{I}}
\newcommand{\cJ}{\mathcal{J}}
\newcommand{\cK}{\mathcal{K}}
\newcommand{\cL}{\mathcal{L}}
\newcommand{\cM}{\mathcal{M}}
\newcommand{\cN}{\mathcal{N}}
\newcommand{\cO}{\mathcal{O}}
\newcommand{\cP}{\mathcal{P}}
\newcommand{\cQ}{\mathcal{Q}}
\newcommand{\cR}{\mathcal{R}}
\newcommand{\cS}{\mathcal{S}}
\newcommand{\cT}{\mathcal{T}}
\newcommand{\cU}{\mathcal{U}}
\newcommand{\cV}{\mathcal{V}}
\newcommand{\cW}{\mathcal{W}}
\newcommand{\cX}{\mathcal{X}}
\newcommand{\cY}{\mathcal{Y}}
\newcommand{\cZ}{\mathcal{Z}}

\newcommand{\vA}{\mathbf{A}}
\newcommand{\vB}{\mathbf{B}}
\newcommand{\vC}{\mathbf{C}}
\newcommand{\vD}{\mathbf{D}}
\newcommand{\vE}{\mathbf{E}}
\newcommand{\vF}{\mathbf{F}}
\newcommand{\vG}{\mathbf{G}}
\newcommand{\vH}{\mathbf{H}}
\newcommand{\vI}{\mathbf{I}}
\newcommand{\vJ}{\mathbf{J}}
\newcommand{\vK}{\mathbf{K}}
\newcommand{\vL}{\mathbf{L}}
\newcommand{\vM}{\mathbf{M}}
\newcommand{\vN}{\mathbf{N}}
\newcommand{\vO}{\mathbf{O}}
\newcommand{\vP}{\mathbf{P}}
\newcommand{\vQ}{\mathbf{Q}}
\newcommand{\vR}{\mathbf{R}}
\newcommand{\vS}{\mathbf{S}}
\newcommand{\vT}{\mathbf{T}}
\newcommand{\vU}{\mathbf{U}}
\newcommand{\vV}{\mathbf{V}}
\newcommand{\vW}{\mathbf{W}}
\newcommand{\vX}{\mathbf{X}}
\newcommand{\vY}{\mathbf{Y}}
\newcommand{\vZ}{\mathbf{Z}}

\newcommand{\va}{\mathbf{a}}
\newcommand{\vb}{\mathbf{b}}
\newcommand{\vc}{\mathbf{c}}
\newcommand{\vd}{\mathbf{d}}
\newcommand{\ve}{\mathbf{e}}
\newcommand{\vf}{\mathbf{f}}
\newcommand{\vg}{\mathbf{g}}
\newcommand{\vh}{\mathbf{h}}
\newcommand{\vi}{\mathbf{i}}
\newcommand{\vj}{\mathbf{j}}
\newcommand{\vk}{\mathbf{k}}
\newcommand{\vl}{\mathbf{l}}
\newcommand{\vm}{\mathbf{m}}
\newcommand{\vn}{\mathbf{n}}
\newcommand{\vo}{\mathbf{o}}
\newcommand{\vp}{\mathbf{p}}
\newcommand{\vq}{\mathbf{q}}
\newcommand{\vr}{\mathbf{r}}
\newcommand{\Vs}{\mathbf{s}}
\newcommand{\vt}{\mathbf{t}}
\newcommand{\vu}{\mathbf{u}}
\newcommand{\vv}{\mathbf{v}}
\newcommand{\vw}{\mathbf{w}}
\newcommand{\vx}{\mathbf{x}}
\newcommand{\vy}{\mathbf{y}}
\newcommand{\vz}{\mathbf{z}}

\newcommand{\vone}{\mathbf{1}}
\newcommand{\vzero}{\mathbf{0}}

\newcommand{\valpha}{{\boldsymbol{\alpha}}}
\newcommand{\vbeta}{{\boldsymbol{\beta}}}
\newcommand{\vgamma}{{\boldsymbol{\gamma}}}
\newcommand{\vdelta}{{\boldsymbol{\delta}}}
\newcommand{\vepsilon}{{\boldsymbol{\epsilon}}}
\newcommand{\vzeta}{{\boldsymbol{\zeta}}}
\newcommand{\veta}{{\boldsymbol{\eta}}}
\newcommand{\vtheta}{{\boldsymbol{\theta}}}
\newcommand{\viota}{{\boldsymbol{\iota}}}
\newcommand{\vkappa}{{\boldsymbol{\kappa}}}
\newcommand{\vlambda}{{\boldsymbol{\lambda}}}
\newcommand{\vmu}{{\boldsymbol{\mu}}}
\newcommand{\vnu}{{\boldsymbol{\nu}}}
\newcommand{\vxi}{{\boldsymbol{\xi}}}
\newcommand{\vomikron}{{\boldsymbol{\omikron}}}
\newcommand{\vpi}{{\boldsymbol{\pi}}}
\newcommand{\vrho}{{\boldsymbol{\rho}}}
\newcommand{\vsigma}{{\boldsymbol{\sigma}}}
\newcommand{\vtau}{{\boldsymbol{\tau}}}
\newcommand{\vupsilon}{{\boldsymbol{\upsilon}}}
\newcommand{\vphi}{{\boldsymbol{\phi}}}
\newcommand{\vchi}{{\boldsymbol{\chi}}}
\newcommand{\vpsi}{{\boldsymbol{\psi}}}
\newcommand{\vomega}{{\boldsymbol{\omega}}}

\newcommand{\rLambda}{\mathrm{\Lambda}}
\newcommand{\rSigma}{\mathrm{\Sigma}}

%--------------------------------------------------------------------
% Add a period to the end of an abbreviation unless there's one
% already, then \xspace.
\makeatletter
\DeclareRobustCommand\onedot{\futurelet\@let@token\@onedot}
\def\@onedot{\ifx\@let@token.\else.\null\fi\xspace}
\def\eg{\emph{e.g}\onedot} \def\Eg{\emph{E.g}\onedot}
\def\ie{\emph{i.e}\onedot} \def\Ie{\emph{I.e}\onedot}
\def\vs{\emph{vs\onedot}}
\def\cf{\emph{cf}\onedot} \def\Cf{\emph{C.f}\onedot}
\def\etc{\emph{etc}\onedot} \def\vs{\emph{vs}\onedot}
\def\wrt{w.r.t\onedot} \def\dof{d.o.f\onedot}
\def\etal{\emph{et al}\onedot}
\makeatother

\title{Towards Universal Image Embeddings: \\ A Large-Scale Dataset and Challenge for Generic Image Representations}

\author{Nikolaos-Antonios Ypsilantis$^1$
\and
Kaifeng Chen$^2$
\and
Bingyi Cao$^2$
\and
Mário Lipovsk\'{y}$^2$
\and
Pelin Dogan-Schönberger$^2$
\and
Grzegorz Makosa$^2$
\and
Boris Bluntschli$^2$
\and
Mojtaba Seyedhosseini$^2$
\and
Ond\v{r}ej Chum$^1$
\and
André Araujo$^2$ \and \\
\vspace{-5pt}
$^1$VRG, FEE, Czech Technical University in Prague \hspace{25pt} $^2$Google
}

\maketitle
\ificcvfinal\thispagestyle{empty}\fi

\begin{abstract}

Fine-grained and instance-level recognition methods are commonly trained and evaluated on specific domains, in a model per domain scenario. Such an approach, however, is impractical in real large-scale applications. In this work, we address the problem of \textbf{universal image embedding}, where a single universal model is trained and used in multiple domains.
First, we leverage existing domain-specific datasets to carefully construct a new large-scale public benchmark for the evaluation of universal image embeddings, with 241k query images, 1.4M index images and 2.8M training images across 8 different domains and 349k classes.
We define suitable metrics, training and evaluation protocols to foster future research in this area.
Second, we provide a comprehensive experimental evaluation on the new dataset, demonstrating that existing approaches and simplistic extensions lead to worse performance than an assembly of models trained for each domain separately.
Finally, we conducted a public research competition on this topic, leveraging industrial datasets, which attracted the participation of more than 1k teams worldwide.
This exercise generated many interesting research ideas and findings which we present in detail.
Project webpage: \url{https://cmp.felk.cvut.cz/univ_emb/}

\end{abstract}

\section{Introduction}
\label{sec:intro}

\begin{figure}[t]
\begin{center}
   \includegraphics[width=1.0\linewidth]{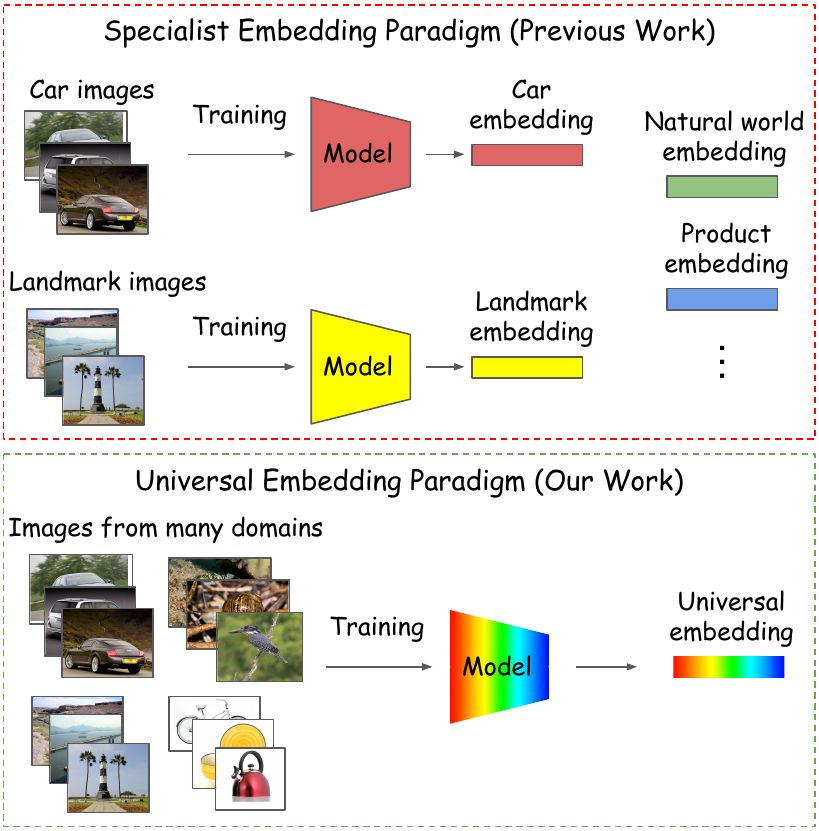}
\end{center}
\vspace{-10pt}
\caption{
   Previous work (top) on learning image embeddings has mainly focused on representations that are specialized for narrow domains, such as cars, landmarks, natural world, products, among others.
   While this may lead to high performance, it cannot scale to meet demands of modern general purpose visual search systems, which are required to identify objects in many domains.
   In this work, we consider the problem of learning universal image embeddings (bottom), which are representations that can encode fine-grained visual information about multiple domains.
   We propose the first large-scale dataset for research on universal image embeddings and additionally present results from a public industrial challenge in this area.
   \vspace{-15pt}
   }
\label{fig:key_fig}
\vspace{-10pt}
\end{figure}

The past decade has witnessed significant progress in image representations that are capable of discriminating objects at a fine-grained or instance level.
\blfootnote{$^*$Correspondence: ypsilnik@fel.cvut.cz, francischen@google.com.}
Several techniques \cite{schroff2015facenet,qian2015fine,song2016deep,chen2022deep} have been demonstrated to learn such image embeddings when given data from a specific domain, for example images of different birds or images of different landmarks.
Recently, there has been growing interest in general purpose visual search systems that can identify objects from many domains \cite{zhai2019learning,bixby,googlelens}.
The use of per-domain models in general-purpose systems is very expensive and generally impractical, since a large number of models would need to be developed and maintained.
The holy grail for this kind of application is a unified model that can discriminate fine-grained objects across several domains, which we refer to as an \emph{universal image embedding}, as per \figref{fig:key_fig}.
Such an universal embedding is a challenging goal as different domains provide different visual cues that are essential for fine-grained and instance-level recognition.
It can be seen as an evolution of generalization in training: generalization beyond training instances in classification with fixed classes, beyond training classes within one domain in an open set problem (such as landmark retrieval), and finally, beyond training classes in multiple domains in the universal embedding problem.
We believe that the field of image embedding learning needs to continue moving forward by considering universal representations as a critical direction of future work.

The main reason holding back research explorations on universal embeddings is the lack of a standard, large-scale dataset -- only small datasets have so far been proposed \cite{wang2015instre,schall2021gpr1200}, or related medium ones that have been constructed with different objectives \cite{almazan2022granularity}.
Today, there are no established strategies to train such models, and their effectiveness on industrial applications is not well-studied either.
We set out to bridge this gap by introducing the following contributions.

\noindent\textbf{Contributions.}
(1) The first large-scale dataset for research on universal image embeddings, referred to as Universal Embedding Dataset (UnED).
The dataset contains more than $4M$ images from $349k$ classes in $8$ different domains, representing diverse \& real use cases: food, cars, online products, clothing, natural world, artworks, landmarks and retail products. 
We leverage already-existing public datasets to construct UnED, carefully combining them into a common format, with standard splits and metrics.
(2) A comprehensive benchmarking and reference implementations of models for research in this area, 
highlighting that specialized models on average outperform universal models trained with simple strategies; nevertheless, the universal models achieve promising results and pave the way for further improvements.
(3) The first 
public competition in this area, the Google Universal Image Embedding Challenge\footnote{\url{https://www.kaggle.com/competitions/google-universal-image-embedding}}, focusing on industrial applications, which attracted more than $1k$ researchers and $21k$ submissions in total.
We report learned lessons from this challenge, which helped open up new research directions.

\section{Related Work}
\label{sec:rw}

\noindent\textbf{Image embedding research.}
Traditionally, academic research on image embedding learning has been conducted with a focus on models which are specialized for a given domain, \ie, a specific object type (\eg, birds, cars, landmarks, etc).
Generally, researchers propose embedding learning techniques which are applied to different domains separately, rather than developing (universal) embedding models which could be applied to all domains combined.
There are three main computer vision sub-communities working in this area, and we review their work in the following paragraphs.

(1) \textit{Deep Metric Learning} -- generally focused on the domains of cars \cite{krause20133d}, products \cite{song2016deep} and birds \cite{wah2011cub}.
Recent papers focus on improved benchmark methodology \cite{musgrave2020reality}, leveraging intra-batch relations \cite{seidenschwarz2021learning}, enhanced sampling \cite{levi2021rethinking} and integrating language guidance \cite{roth2022integrating}.

(2) \textit{Instance-level Retrieval} -- generally focused on the domain of landmarks \cite{weyand2020google,radenovic2018revisiting}.
Recent work reports improvements to models \cite{cao2020unifying,yang2021dolg} and re-ranking strategies \cite{tan2021instance,lee2022correlation}.
A recent survey can be found in \cite{chen2022deep}.

(3) \textit{Person Recognition/Re-Identification} -- focusing on person-related data such as face \cite{kemelmacher2016megaface,maze2018ijbc} or full-body \cite{li2014deepreid,zheng2015scalable}.
Recent research introduces quality-adaptive margins \cite{kim2022adaface}, joint optimization of data/architecture/loss \cite{zhang2022towards}, cross-domain learning \cite{zhang2022adaptive} and improved pre-training \cite{zhu2022pass}.

\begin{table}[t]
\footnotesize
\centering
{
\begin{tabularx}{0.89\textwidth}{l c c c c}
Dataset  & Year & \# Images & \# Domains & \# Classes  \\
\midrule
INSTRE \cite{wang2015instre} & 2015 & $28k$ & $3$ & $250$  \\
GPR1200 \cite{schall2021gpr1200} & 2021 & $12k$ & $6$ & $1.2k$  \\
MRT \cite{almazan2022granularity} & 2022 & $267k$ & $6$ & $23k$  \\
\midrule
UnED (ours) & 2023 & \bf{4.1M} & \bf{8} & \bf{349k} \\
\end{tabularx}
}
\vspace{-5pt}
\caption{Comparison of our dataset against existing ones. Our dataset is significantly larger, with one order of magnitude more images and classes.
\vspace{-10pt}
}
\label{tab:rw_datasets}
\end{table}

\noindent\textbf{Universal embedding datasets.}
To the best of our knowledge, no truly large-scale datasets for unified embedding model evaluation exist.
\tabref{tab:rw_datasets} compares our new dataset against the three existing related datasets we are aware of.
INSTRE \cite{wang2015instre} contains $1k$ query images and $27k$ index images of $250$ classes, covering $3$ domains: landmarks, planar objects and other daily objects.
More similar in spirit to our work, GPR1200 \cite{schall2021gpr1200} introduces an evaluation set containing $12k$ images in total (from $1.2k$ classes), constructed by collecting images from existing public collections in $6$ domains: landmarks, sketches, natural world, products, planar objects and faces.
The recent MRT \cite{almazan2022granularity} focuses on adapting pre-trained models using unlabeled data from $6$ different domains, also reusing images from existing collections: aircrafts, cars, birds, flowers, food and products.
In its standard setup, MRT's training set discards class labels to address how well models are able to adapt in the absence of supervision, but the same splits could potentially be reused in a supervised setup.
MRT's evaluation is performed for each domain separately.

Our newly-introduced benchmark differs substantially from these three, most notably on the scale aspect, comprising images from $8$ domains: food, cars, online products, clothing, natural world, artworks, landmarks and retail products.
With $241,986$ query images, $1,397,126$ index images and $2,831,222$ training images, we provide $15\times$ the number of images and $15\times$ the number of classes compared to previous datasets\footnote{In the context of this work, the term class is used interchangeably for either fine-grained categories or instances, since the proposed benchmark consists of fine categories at levels of granularity which make the most sense for each domain.}.
Differently from INSTRE and GPR1200, we provide a large training set, enabling researchers to directly compare different models under the same standard training data.
In contrast to MRT, we provide labels in the training set to allow models to better adapt to the domains of interest -- in our experience, this setting is more common in practice.
Additionally, we consider the more challenging evaluation setup where the index contains images from all domains, requiring the model to distinguish images from multiple domains simultaneously.
Finally, we would like to emphasize that the public research competition we conducted was the first to assess performance of universal image embeddings; previous embedding learning challenges focused only on per-domain evaluations \cite{gld21challenge,deepfashion2challenge,ebaychallenge,aliproductschallenge}.

\begin{table}[t]
  \centering
  \scalebox{0.77}{
\begin{tabular}{cccc}
\toprule
Dataset & Domain & Train images & Train classes\\
\midrule
Food2k~\cite{min2023large} & food & 472,349 & 900 \\
CARS196~\cite{krause20133d}  & cars & 6,346 & 78\\
SOP~\cite{song2016deep} &  online products & 48,942 & 9,054 \\
InShop~\cite{liu2016deepfashion} & clothing & 20,897 & 3,198 \\
iNaturalist (2018)~\cite{van2018inaturalist}  & natural world & 273,929 & 4,552 \\
Met~\cite{ypsilantis2021met}  & artworks & 397,121 & 224,408  \\
GLDv2~\cite{weyand2020google} & landmarks & 1,422,914 & 73,182 \\
Rp2k~\cite{peng2020rp2k} & retail products & 188,724 & 1,074 \\
\hdashline
Total & All & 2,831,222 & 316,446 \\
\bottomrule
\end{tabular}
}

  \vspace{-7pt}
  \caption{
  The different domains the UnED subsets span, along with the training splits and the corresponding training classes.
  \label{tab:domains_train}
  \vspace{-10pt}
  }
\end{table}

\noindent\textbf{Universal embedding techniques} have not been thoroughly investigated in previous work.
An early attempt by Feng \etal \cite{feng2020unifying} used data from five domains in a distillation approach to combine the knowledge of specialized models, three at a time, into a universal one.
Our effort differs from theirs as we introduce a much larger dataset where a significant number of domains must be jointly considered.
A more recent method relevant to this problem is Grappa \cite{almazan2022granularity}, which aims to adapt a pre-trained model using unlabeled data from several domains combined -- their setup is different from ours since we consider supervision to be available for the different domains in our task.

\noindent\textbf{Other relevant literature.}
\cite{berman2019multigrain} propose to learn a unified embedding for recognition at different levels of granularity; models are trained on ImageNet with both classification and ranking losses, and testing is done both on ImageNet and instance-level retrieval datasets. 
\cite{hu2015deep} propose metric learning knowledge transfer between datasets, but those are within the same broad modality of person recognition. 
The OmniBenchmark dataset \cite{zhang2022omnibenchmark} aims to unify image representation efforts for classification tasks, gathering $1M$ images from $7k$ classes and $21$ domains.
Evaluation in OmniBenchmark is per-domain, with linear classifier probing.
\section{The Universal Embedding Dataset}
\label{sec:dataset}

The Universal Embedding Dataset (UnED) contains images that come from several publicly available datasets and benchmarks used to evaluate the performance in various tasks, including image retrieval, instance-level and fine-grained recognition. 
The new dataset covers popular visual domains listed in Table~\ref{tab:domains_train}.
These domains were selected to simulate the environment of universal recognition application, and each of them already has its own commercial applications~\cite{art,products,food,natural,landmarks,cars}.
Visual cues used for fine-grained or instance level recognition in the selected domains differ substantially, which pronounces the need for a universal embedding approach and makes the dataset challenging.
At the same time, none of the domains addresses or benefits from methods exploiting any type of biometric information (such as faces), that would allow for human identification.

\begin{table*}[h]
  \centering
  \scalebox{0.66}{
\begin{tabular}{c|ccccc|ccccc}
\hline
& \multicolumn{5}{c|}{Val split} & \multicolumn{5}{c}{Test split} \\
\hline
Dataset & query images & query classes & index images & index classes & unseen classes & query images & query classes & index images & index classes & unseen classes \\
\hline
Food2k & 49,323 & 100 & $\star$ & $\star$ & \cmark &  9,979 & 1,000 & $\star$ & $\star$ & \cmark  \\
CARS196  & 1,708 & 20 & $\star$ & $\star$ & \cmark & 8,131 & 98 & $\star$ & $\star$ & \cmark \\
SOP  & 10,609 & 2,264 & $\star$ & $\star$ & \cmark & 60,502 & 11,316 & $\star$ & $\star$ & \cmark  \\
InShop  & 4,985 & 799 & $\star$ & $\star$ & \cmark  & 14,218 & 3,985 & 12,612 & 3,985& \cmark \\
iNaturalist (2018) & 51,917& 1,138 & $\star$ & $\star$ & \cmark  & 136,093 & 2,452 & $\star$ & $\star$& \cmark \\
Met & 129 & 111 & 38,307 & 33,501  & \xmark &  1,003 & 734& 397,121 & 224,408 & \xmark \\
GLDv2 & 157,556 & 8,131 & $\star$ & $\star$ & \cmark & 1129 & 318  & 761,757 & 101,302& \cmark$\dagger$ \\
Rp2k & 17,185 & 120 & $\star$ & $\star$ & \cmark & 10,931 & 1,186 & $\star$ & $\star$ & \cmark \\
\hdashline
Total & 293,412 & 12,683 & 331,590 & 46,073 & - & 241,986 & 21,089 & 1,397,126 & 345,747 & -\\
\hline
\end{tabular}
}
  \vspace{-7pt}
  \caption{Query and index subsets for the validation and test sets of each subset of UnED.
  If index images are missing for a specific subset, it means that for this dataset queries are used as index as well.
  $\star$ means that the index set statistic matches the corresponding query set statistic for the same dataset split.
  $\dagger$ means that some classes are seen during training, but not all of them.
  \label{tab:query_index}
  \vspace{-10pt}
  }
\end{table*}
\subsection{Datasets and splits details}
Each of the datasets comes with predefined training and testing splits, some of them, but not all, also provide a validation set.
For the sake of fair comparison in the future, for each of the domains (datasets) we define train, validation and test sets. These sets do not necessarily exactly correspond to the original splits.
The specific statistics for the training splits are given in Table~\ref{tab:domains_train}, while those for the validation and test splits are given in Table~\ref{tab:query_index}.
The train and validation sets are used for training and hyperparameter tuning, while the test sets are used for evaluation. 
The test and validation sets are further divided into two subsets. The \emph{index set} contains database images to be retrieved, while the \emph{query set} are images that serve to query by example, details are summarized in Table~\ref{tab:query_index}.  The index and query set splits in validation are provided in order to allow a retrieval validation to have a better proxy for the final task.
For some domains, the query and index sets are identical. This is indicated by the $\star$ symbol in Table~\ref{tab:query_index}. In such cases, the query itself is always excluded from the retrieved results.

The \textbf{food domain} is represented by the recently published Food2k dataset~\cite{min2023large}, the largest to date fine-grained food classification dataset.
New training and testing splits are defined so that there are disjoint train and test classes. Further, for the test classes, the image instances are subsampled (including near-duplicate removal) to have 
no more than 10 images per class. Such a reduction is conducted to keep the dataset challenging and to avoid saturated scores (as many embeddings would obtain high perfomance in the presence of near-duplicated images).
For the \textbf{car domain}, the Stanford Cars~\cite{krause20133d} (known as CARS196) is used. This dataset is widely used for the task of metric learning and image retrieval.
The standard training set is split into training and validation sets (78 and 20 categories respectively), while the original test set of 98 categories is preserved.
The Stanford Online Products dataset~\cite{song2016deep} is leveraged as the \textbf{online product domain}. The original training set is split into training and validation sets, the test set remains as defined in~\cite{song2016deep}.
For the \textbf{domain of clothing}, the InShop retrieval part of the DeepFashion dataset~\cite{liu2016deepfashion} is used.
The original training set is divided into train and validation, while the test set is kept identical, including the separate query and index sets.
The \textbf{natural world domain} is represented by the large-scale iNaturalist dataset (2018 version)~\cite{van2018inaturalist}. The retrieval splits of~\cite{brown2020smooth} are adopted,
the training set is further divided into train and validation.
The \textbf{artwork domain} is covered by the Met dataset~\cite{ypsilantis2021met}. 
In this domain, the test classes are a subset of the train ones. 
The training set of catalog images of artworks is used as an index set, while queries are given as user photographs taken in the exhibition by mobile devices. 
For the Met dataset, all the original splits are preserved.
The GLDv2 Dataset~\cite{weyand2020google} serves as a source for the \textbf{landmark domain}. The train-clean version of the dataset is used to provide the train and the validation sets.
The public and private queries are merged to define the query part of the test set, while the original index set is used as the index part of the test set. 
The last \textbf{domain of retail products} is given by the Rp2k dataset~\cite{peng2020rp2k}, originally designed for image classification. Similarly to the Food2k dataset, new splits are created to have disjoint training and test classes. 
The test classes are also reduced to up to 10 instances per class.

\subsection{Evaluation protocol}

Each image is described by a 64-dimensional embedding.
Low dimensionality is a crucial factor for practical large-scale applications, which is a natural target for fine-grained recognition of many classes from a number of different domains.
The index contains embeddings of index images from \textbf{all} domains. 
It is not allowed to exploit the information about the query and/or result domains.
The evaluation is performed by Euclidean-distance retrieval between the query embedding and the embeddings of images in the index.
\textbf{The embedding of the query is compared against the embeddings of the merged index set}, producing a ranked list of images.
The first metric used to quantify the quality of this ranked list is 
the commonly used 
Recall@1 $(R@1)$, which is equivalent to the nearest neighbour accuracy.
For a given query, it only takes into account the predicted top neighbor, being equal to 1 if it comes from the same class as the query and 0 if it doesn't, and finally it is averaged across all queries $Q$.
Mathematically, it is defined as:
\begin{equation} \label{eq:2}
    R@1=\frac{1}{Q} \sum_{q=1}^Q r e l_q(1),
\end{equation}
where $r e l_q(j)$ denotes the relevance of image at rank $j$ for query $q$ (binary indicator).
Additionally, the precision of the top-5 neighbors list is calculated, \ie how many of the 5 neighbors come from the same class. 
If the number of index images of the same class ($n_q$) as the query are less than 5, the precision at $n_q$ is calculated instead.
This metric averaged across all queries $Q$ is called modified Mean Precision at 5 (mMP@5) and has the following definition:
\begin{equation} \label{eq:3}
    mMP @ 5=\frac{1}{Q} \sum_{q=1}^Q \frac{1}{\min \left(n_q, 5\right)} \sum_{j=1}^{\min \left(n_q, 5\right)} r e l_q(j),
\end{equation}
where $j$ is the index of the neighbors of $q$, sorted in descending order by their similarity.
In the case of a query that has multiple classes assigned to it (as for some GLDv2 queries), each of them is considered a correct prediction.

Let us also highlight that, in contrast to some image embedding benchmarks \cite{radenovic2018revisiting,weyand2020google,musgrave2020reality}, we do not include mean Average Precision (mAP) as one of our core evaluation metrics.
We find that mAP has many drawbacks, for example, capturing differences in scores even for changes in sorting of low-ranked positions that in practice do not matter.
Besides, mAP is unintuitive, being difficult to interpret exactly what a given value means (the AP meaning changes for each query, depending on the number of expected results for each).
For these reasons, we find it more suitable in our case to rely on simple and practical metrics such as the above, which are easily interpretable and capture well the desired system behavior: rank relevant images high, and focus mainly on the very top positions.
Despite this, recognizing that the community still relies on mAP in many cases, we include results using it in the Appendix.
\begin{figure}[ht]
\begin{center}
\includegraphics[width=1.0\linewidth]{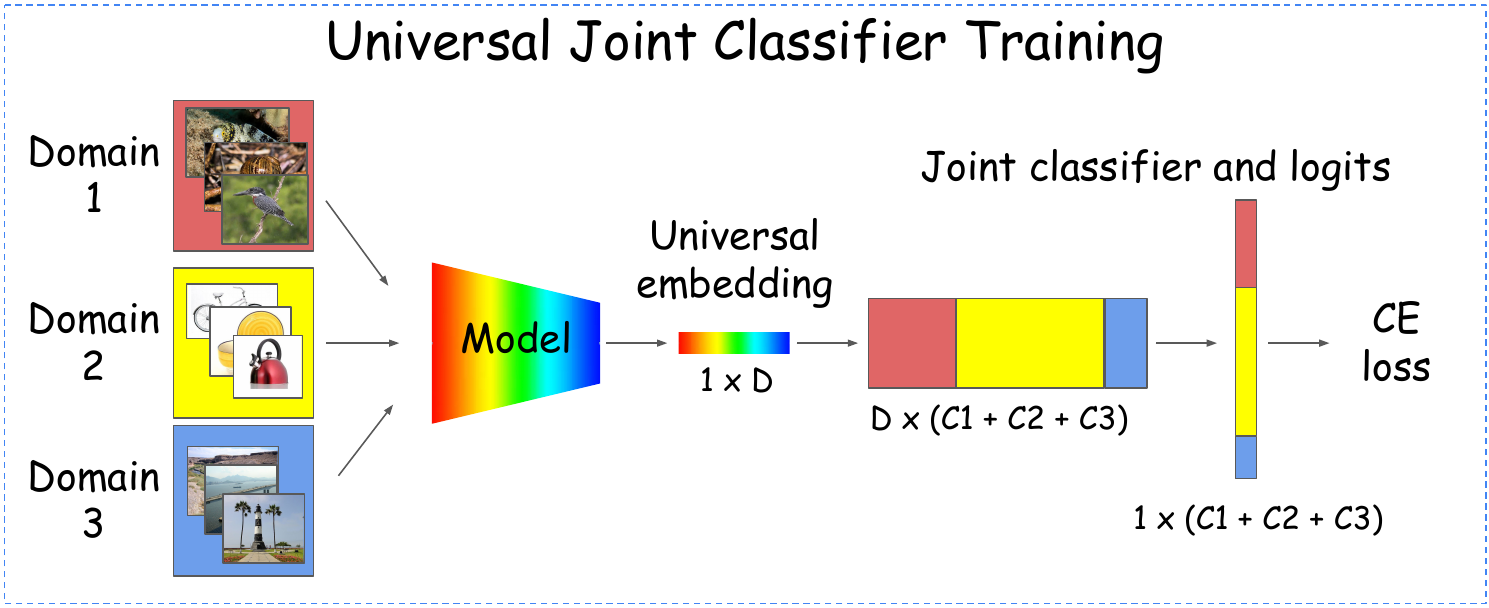}
\includegraphics[width=1.0\linewidth]{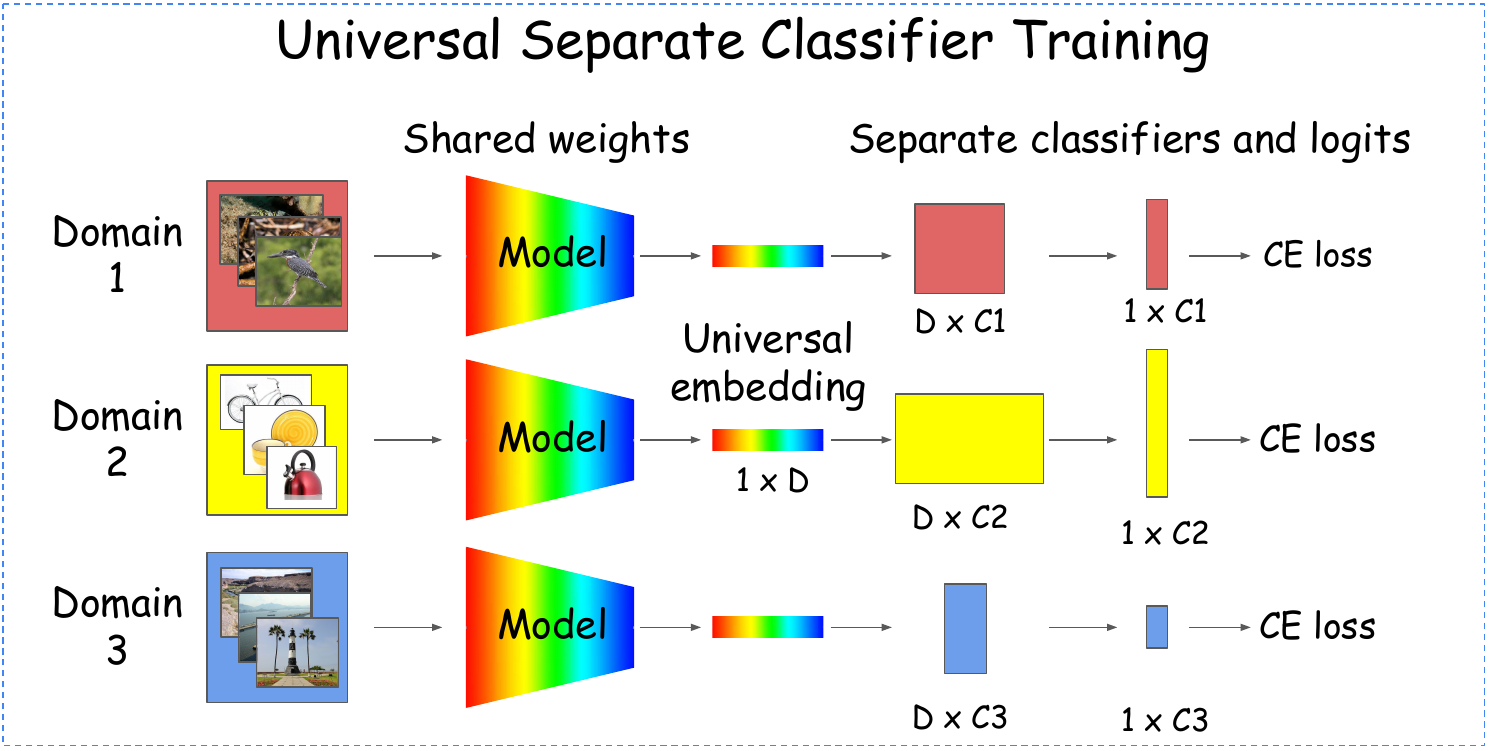}
\end{center}
\vspace{-18pt}
   \caption{Universal embedding training methods used as baselines in this work. In joint classifier training (top), a single classifier head on top of the backbone is trained on all combined classes. In separate classifier training (bottom), classifier heads for different domains are trained separately, given a shared backbone.
   For the sake of compactness, only 3 of the 8 domains are presented.
   \vspace{-8pt}
   }
\label{fig:baselines}
\end{figure}

\section{Benchmarking}
\label{sec:benchmarking}
In this section we describe the baseline approaches that are evaluated on the proposed benchmark, in order to offer a testbed for future comparisons.

\subsection{Baseline approaches}

\noindent\textbf{Pretrained models.}
First, different pretrained models are evaluated by extracting off-the-shelf embeddings.
For this evaluation, the original dimensionality is used.
More specifically, we benchmark standard ImageNet AugReg pretraining (IN)~\cite{steiner2021train}, image-text foundation model OpenAI CLIP~\cite{radford2021learning} and recent DINOv2~\cite{oquab2023dinov2} that has shown to produce very strong generic features, all using a ViT-B/16 backbone (768-D).
We also benchmark MultiGrain~\cite{berman2019multigrain} embeddings from a ResNet50 backbone (2048-D), as they have been trained for a relevant task.
For the ones that utilise the ViT backbone, the [CLS] token of the last layer is used as the embedding; for the Multigrain, the pooled representantion before the classifier.
In both cases they undergo \l2 normalization.
For the IN and CLIP models which are later used as initialization for finetuning, we perform PCA-Whitening~\cite{jegou2012negative} to additionally reduce the dimensionality to 64-D, trained on the union of a subset of random training images from each domain.
\begin{table*}[ht]
  \centering
  \definecolor{LightCyan}{rgb}{0.88,1,1}
\definecolor{LightRed}{rgb}{1.0,0.9,0.9}
\definecolor{LightYellow}{rgb}{1.0,1.0,0.7}
\scalebox{0.526}{
\begin{tabular}{c|cc|cc|cc|cc|cc|cc|cc|cc|cc}
\hline
& \multicolumn{2}{c|}{Food2k} & \multicolumn{2}{c|}{CARS196}& \multicolumn{2}{c|}{SOP}&\multicolumn{2}{c|}{InShop}&\multicolumn{2}{c|}{iNat}& \multicolumn{2}{c|}{Met}& \multicolumn{2}{c|}{GLDv2}& \multicolumn{2}{c|}{Rp2k}&  \multicolumn{2}{c}{Mean}\\ 
\hline
 Model & mMP@5 & R@1 & mMP@5 & R@1 & mMP@5 & R@1 & mMP@5 & R@1 & mMP@5 & R@1 & mMP@5 & R@1 & mMP@5 & R@1 & mMP@5 & R@1 & mMP@5 & R@1\\
\hline
\multicolumn{19}{c}{\textbf{Off-the-shelf}}\\
\hline
IN (\textbf{\red{768-D}})
& 31.1
& 44.1
& 41.4
& 54.1
& 43.7
& 65.6
& 35.5
& 53.9
& 67.1
& 74.2
& 21.1
& 30.8
& 14.8
& 25.2
& 52.9
& 74.3
& 38.4
& 52.8
\\
CLIP (\textbf{\red{768-D}})
& 29.4
& 42.9
& 74.7
& 82.2
& 44.2
& 65.4
& 37.2
& 56.0
& 52.4
& 61.9
& 21.4
& 28.5
& 20.4
& 31.0
& 38.6
& 59.9
& 39.8
& 53.5
\\
DINOv2 (\textbf{\red{768-D}})
& 39.9 
& 51.4
& 67.1 
& 79.5
& 35.6 
& 56.0
& 17.4
& 33.4
& 71.2
& 77.6
& 38.3
& 48.1
& 35.4
& 51.7
& 46.6
& 67.8
& 43.9
& 58.2
\\
MultiGrain (\textbf{\red{2048-D}})
& 14.8
& 24.2
& 27.1
& 39.7
& 37.7
& 59.5
& 19.8
& 34.3
& 34.9
& 43.4
& 12.0
& 16.1
& 8.6
& 16.7
& 51.5
& 73.3
& 25.8
& 38.4
\\
\hdashline
IN+PCAw 
& 19.1
& 29.1
& 29.0
& 37.8
& 30.5
& 51.2
& 19.6
& 31.6
& 50.9
& 57.9
& 8.0
& 11.0
& 8.3
& 13.2
& 37.6
& 57.8
& 25.4
& 36.2
\\
CLIP+PCAw
& 23.4
& 34.6
& 62.8
& 72.7
& 36.5
& 57.0
& 27.0
& 41.8
& 42.7
& 51.1
& 12.1
& 15.8
& 11.9 
& 17.6 
& 32.0
& 51.8
& 31.0 
& 42.8 
\\
\hline
\multicolumn{19}{c}{\textbf{Specialists}}\\
\hline
IN+Oracle
& 49.9\scriptsize{$\pm$0.5}
& 62.8\scriptsize{$\pm$0.5}
& 61.9\scriptsize{$\pm$0.6}
& 71.8\scriptsize{$\pm$0.3}
& 60.9\scriptsize{$\pm$0.5}
& 78.1\scriptsize{$\pm$0.3}
& 66.3\scriptsize{$\pm$0.2}
& 85.9\scriptsize{$\pm$0.2}
& \textbf{70.1\scriptsize{$\pm$0.1}}
& \textbf{75.2\scriptsize{$\pm$0.1}}
& 20.4\scriptsize{$\pm$1.0}
& 24.9\scriptsize{$\pm$0.7}
& 31.2\scriptsize{$\pm$0.5}
& 43.1\scriptsize{$\pm$0.5}
& 73.6\scriptsize{$\pm$0.2}
& \textbf{87.1\scriptsize{$\pm$0.1}}
& 54.9\scriptsize{$\pm$0.4}
& 66.6\scriptsize{$\pm$0.3}
\\
CLIP+Oracle 
& \textbf{51.5\scriptsize{$\pm$0.3}}
& \textbf{63.7\scriptsize{$\pm$0.2}}
& \textbf{83.4\scriptsize{$\pm$0.8}}
& \textbf{88.5\scriptsize{$\pm$0.3}}
& 65.8\scriptsize{$\pm$0.3}
& 81.2\scriptsize{$\pm$0.2}
& 68.0\scriptsize{$\pm$0.3}
& 86.2\scriptsize{$\pm$0.3}
& 67.3\scriptsize{$\pm$0.1}
& 73.0\scriptsize{$\pm$0.2}
& \textbf{27.6\scriptsize{$\pm$1.4}}
& \textbf{32.9\scriptsize{$\pm$1.5}}
& \textbf{35.1\scriptsize{$\pm$0.7}} 
& \textbf{46.6\scriptsize{$\pm$0.4}}
& 69.7\scriptsize{$\pm$0.2}
& 84.4\scriptsize{$\pm$0.2}
& \textbf{59.6\scriptsize{$\pm$0.3}}
& \textbf{70.4\scriptsize{$\pm$0.2}}
\\
\hline
\multicolumn{19}{c}{\textbf{Universal models}}\\
\hline
IN+UJCDS 
& 49.1\scriptsize{$\pm$1.7} 
& 60.6\scriptsize{$\pm$1.2}
& 50.9\scriptsize{$\pm$1.7}
& 60.9\scriptsize{$\pm$1.9}
& 56.9\scriptsize{$\pm$1.4}
& 74.4\scriptsize{$\pm$0.9}
& 60.3\scriptsize{$\pm$2.8}
& 78.0\scriptsize{$\pm$2.1}
& 67.6\scriptsize{$\pm$0.7}
& 72.5\scriptsize{$\pm$0.8}
& 3.8\scriptsize{$\pm$0.3}
& 4.6\scriptsize{$\pm$0.4}
& 29.3\scriptsize{$\pm$1.2}
& 42.0\scriptsize{$\pm$1.3}
& 71.6\scriptsize{$\pm$2.3}
& 85.1\scriptsize{$\pm$1.4}
& 48.7\scriptsize{$\pm$0.3}
& 59.8\scriptsize{$\pm$0.3}
\\
CLIP+UJCDS 
& \blue{50.9\scriptsize{$\pm$0.6}}
& \blue{62.2\scriptsize{$\pm$0.7}}
& 77.4\scriptsize{$\pm$1.1}
& 83.6\scriptsize{$\pm$0.9}
& 59.1\scriptsize{$\pm$0.3}
& 75.8\scriptsize{$\pm$0.2}
& 62.6\scriptsize{$\pm$0.7}
& 80.3\scriptsize{$\pm$0.6}
& 64.2\scriptsize{$\pm$0.1}
& 70.0\scriptsize{$\pm$0.1}
& 2.5\scriptsize{$\pm$0.1}
& 3.3\scriptsize{$\pm$0.3}
& \blue{32.8\scriptsize{$\pm$0.4}}
& \blue{44.7\scriptsize{$\pm$0.6}}
& 69.6\scriptsize{$\pm$0.4}
& 83.8\scriptsize{$\pm$0.3}
& 52.4\scriptsize{$\pm$0.3}
& 63.0\scriptsize{$\pm$0.2}
\\
\hdashline
IN+UJCRR 
& 48.6\scriptsize{$\pm$0.2}
& 60.3\scriptsize{$\pm$0.3}
& 62.9\scriptsize{$\pm$0.2}
& 71.3\scriptsize{$\pm$0.4}
& 64.7\scriptsize{$\pm$0.3}
& 80.2\scriptsize{$\pm$0.3}
& 74.0\scriptsize{$\pm$0.4}
& 89.9\scriptsize{$\pm$0.2}
& 68.3\scriptsize{$\pm$0.1}
& 73.3\scriptsize{$\pm$0.1}
& 5.5\scriptsize{$\pm$0.2}
& 7.0\scriptsize{$\pm$0.6}
& 21.1\scriptsize{$\pm$0.2}
& 31.6\scriptsize{$\pm$0.3}
& \boldblue{74.1\scriptsize{$\pm$0.3}}
& 86.8\scriptsize{$\pm$0.1}
& 52.4\scriptsize{$\pm$0.1}
& 62.6\scriptsize{$\pm$0.1}
\\
CLIP+UJCRR
& 50.1\scriptsize{$\pm$0.1}
& 62.0\scriptsize{$\pm$0.2}
& \blue{80.0\scriptsize{$\pm$0.2}}
& \blue{85.4\scriptsize{$\pm$0.7}}
& \boldblue{68.6\scriptsize{$\pm$0.2}}
& \boldblue{82.7\scriptsize{$\pm$0.1}}
& \boldblue{77.0\scriptsize{$\pm$0.3}}
& \boldblue{91.1\scriptsize{$\pm$0.1}}
& 63.7\scriptsize{$\pm$0.2}
& 69.5\scriptsize{$\pm$0.3}
& 4.6\scriptsize{$\pm$0.5}
& 5.8\scriptsize{$\pm$0.7}
& 25.5\scriptsize{$\pm$0.3}
& 36.0\scriptsize{$\pm$0.4}
& 70.1\scriptsize{$\pm$0.3}
& 84.1\scriptsize{$\pm$0.2}
& \blue{55.0\scriptsize{$\pm$0.1}}
& 64.6\scriptsize{$\pm$0.1}
\\
\hdashline
IN+USCRR 
& 48.3\scriptsize{$\pm$1.3}
& 60.9\scriptsize{$\pm$0.9}
& 58.9\scriptsize{$\pm$1.3}
& 69.7\scriptsize{$\pm$1.2}
& 61.9\scriptsize{$\pm$0.7}
& 78.7\scriptsize{$\pm$0.4}
& 70.4\scriptsize{$\pm$0.7}
& 88.3\scriptsize{$\pm$0.5}
& \blue{69.1\scriptsize{$\pm$0.1}}
& \blue{74.2\scriptsize{$\pm$0.1}}
& 7.3\scriptsize{$\pm$0.7}
& 9.7\scriptsize{$\pm$0.9}
& 21.3\scriptsize{$\pm$1.0}
& 31.4\scriptsize{$\pm$1.6}
& \boldblue{74.1\scriptsize{$\pm$0.4}}
& \boldblue{87.1\scriptsize{$\pm$0.4}}
& 51.4\scriptsize{$\pm$0.2}
& 62.5\scriptsize{$\pm$0.3}
\\
CLIP+USCRR 
& 49.5\scriptsize{$\pm$0.5}
& 61.4\scriptsize{$\pm$0.3}
& 79.0\scriptsize{$\pm$0.8}
& 84.9\scriptsize{$\pm$0.8}
& 65.6\scriptsize{$\pm$0.3}
& 81.1\scriptsize{$\pm$0.1}
& 73.1\scriptsize{$\pm$0.1}
& 89.4\scriptsize{$\pm$0.1}
& 64.4\scriptsize{$\pm$0.6}
& 70.5\scriptsize{$\pm$0.5}
& 8.6\scriptsize{$\pm$0.2}
& 10.8\scriptsize{$\pm$0.1}
& 25.3\scriptsize{$\pm$0.3}
& 36.5\scriptsize{$\pm$0.2}
& 71.1\scriptsize{$\pm$0.8}
& 85.1\scriptsize{$\pm$0.4}
& 54.6\scriptsize{$\pm$0.4}
& 64.9\scriptsize{$\pm$0.3}
\\
\hdashline
IN+USCSS 
& 49.0\scriptsize{$\pm$0.2}
& 61.7\scriptsize{$\pm$0.2}
& 53.4\scriptsize{$\pm$2.5}
& 64.3\scriptsize{$\pm$2.2}
& 62.0\scriptsize{$\pm$0.5}
& 78.8\scriptsize{$\pm$0.3}
& 67.6\scriptsize{$\pm$0.1}
& 87.2\scriptsize{$\pm$0.1}
& 68.3\scriptsize{$\pm$0.4}
& 73.5\scriptsize{$\pm$0.3}
& 8.4\scriptsize{$\pm$1.2}
& 10.7\scriptsize{$\pm$1.7}
& 28.0\scriptsize{$\pm$0.2}
& 40.6\scriptsize{$\pm$0.1}
& 73.5\scriptsize{$\pm$0.5}
& \boldblue{87.1\scriptsize{$\pm$0.4}}
& 51.3\scriptsize{$\pm$0.3}
& 63.0\scriptsize{$\pm$0.1}
\\
CLIP+USCSS
& 49.8\scriptsize{$\pm$0.7}
& 62.0\scriptsize{$\pm$0.7}
& 76.4\scriptsize{$\pm$2.0}
& 83.4\scriptsize{$\pm$1.4}
& 65.8\scriptsize{$\pm$1.1}
& 81.3\scriptsize{$\pm$0.6}
& 71.0\scriptsize{$\pm$1.2}
& 88.5\scriptsize{$\pm$0.9}
& 65.3\scriptsize{$\pm$1.0}
& 71.4\scriptsize{$\pm$0.8}
& \blue{9.9\scriptsize{$\pm$1.8}}
& \blue{12.7\scriptsize{$\pm$2.0}}
& 31.5\scriptsize{$\pm$1.3}
& 42.8\scriptsize{$\pm$1.9}
& 70.1\scriptsize{$\pm$0.9}
& 84.8\scriptsize{$\pm$0.5}
& \blue{55.0\scriptsize{$\pm$0.9}}
& \blue{65.9\scriptsize{$\pm$0.8}}
\\
\hline
\end{tabular}
}
  \vspace{-5pt}
  \caption{Model evaluation on UnED 
  test set, all results for 64-D (unless stated otherwise) \l2 normalized descriptors.
  PCAw : Projection to 64-D by PCA-Whitening learned on a subset of the UnED training set.
  UJCDS: Universal Joint Classifier Dataset Size sampling,
  UJCRR: Universal Joint Classifier Round Robin sampling,
  USCRR: Universal Separate Classifier Round Robin sampling,
  USCSS: Universal Separate Classifier Specialist Steps sampling,
  For 64-D embeddings, we highlight with:
  \blue{Blue}: best unified model for that domain, \textbf{Bold}: best for that domain across all baselines.
  The evaluation is averaged across (i) queries of each domain separately and (ii) across all domains, \ie balanced average (``Mean'' column) of the UnED test set.
  Note that all queries are compared against the merged index set that contains all domains.
  \label{tab:results}
  \vspace{-10pt}
  }
\end{table*}

\noindent\textbf{Training on UnED.}
The ViT-B/16~\cite{dosovitskiy2020image} backbone initialized with either IN or CLIP is further finetuned on the UnED training set.
In particular, the [CLS] token of the last layer is \l2 normalized and then projected to 64-D using a (trainable) linear layer.
The 64-D embedding is \l2 normalized again, as is common in end-to-end image search architectures that include a projection layer~\cite{gordo2016deep}, forming the embedding used in the search.
For learning the embedding, we use the Softmax Cross Entropy loss (CE) on top of linear layer with no bias and \l2 normalized rows (Normalized Softmax Loss~\cite{zhai2018classification}), which is a commonly used classification based objective in the metric learning literature.

Given that the ultimate goal is to achieve (or even overcome) specialist performance with only one universal embedding, we first train one model on each domain (specialists), in order to get an estimate of the specialist performance that can be achieved on that domain.
Then, we train the universal model on all domains at the same time to examine how far direct generalizations of the specialist training methods are from achieving specialist performance in each specific domain.
\\
\noindent\textit{Specialist embedding training.}
For each domain, a specialist model is trained using only training samples from the particular domain.
The validation set of that domain is used in order to prevent overfitting by early stopping at the epoch that maximizes validation $R@1$.
\\
\noindent\textit{Universal embedding training.}
Universal models are trained on the union of the training sets of the domains, with the total number of training classes being equal to the sum of the training classes of the different domains.
The validation set during universal training consists of the union of the validation sets across all domains, \ie the index set corresponds to the merged index sets and the query set corresponds to the merged query sets.
It is used to perform early stopping at the epoch that maximizes the balanced average $R@1$ across all domains.
We choose this way of performing validation when training universal models as it matches the final evaluation.

We examine two different approaches for the final classification layer of the universal model, \ie a Joint (common) classifier for all classes of the UnED training set, or a Separate classifier for each domain; both are visualized in an example in Figure~\ref{fig:baselines}.
For the latter, we take into account the domain that the training sample comes from, and only forward it through the corresponding classifier to produce the loss.

When training on multiple domains, the sampling strategy of the domains has to be taken into account, as imbalances are inevitable.
The model is trained with batches that contain samples from only one domain at a time, perform an optimisation step after every batch, and we examine the following schemes:
(i) sampling domains with probability that is proportional to their frequency in the training set (Dataset Size (DS) sampling), 
(ii) sampling each next batch in a Round-Robin manner (RR sampling),
in a cyclic order, resulting in a balanced sampling and
(iii), following~\cite{lac+21}, sampling according to the steps needed for the corresponding Specialist to reach maximum performance in its domain (Specialist Steps (SS) sampling).

\noindent \paragraph{Implementation details.}
We use the following standard metric learning training augmentations: resizing the image to $256\times256$ followed by random cropping to $224\times224$, random horizontal flipping with probability of 0.5 and normalization with pretraining statistics of the backbone.
During test time, we resize to 224$\times$224 and normalize with pretraining statistics.
The projection and the classification layers are randomly initialized.
Optimization uses the AdamW algorithm with a batch size of 128 images; the scale parameter of the Normalized Softmax loss is kept fixed at the value of 16 and we use the following learning rate schedule: for the first 2 epochs, only the classifier is trained with a learning rate of $1e-3$, while the embedding is kept fixed, and weight decay is introduced with a value of $1e-6$.
Then, everything is trained until 30 epochs are reached, with the backbone being updated with a learning rate of $1e-5$.
Those hyperparameters are common to all experiments, for both specialists and universal models, chosen by performing the best on average across all domains.
All the experiments (unless deterministic) are run for 3 different seeds; the mean and standard deviation of each experiment is reported.
Training and validation for a universal embedding model takes approximately 16 hours in a Google Cloud TPU v4, using the Scenic framework~\cite{dga+21}.

\subsection{Experimental results and discussion}
The experimental results are summarized in Table~\ref{tab:results}, where we report performance of each baseline both across queries of each domain, as well as the balanced average of all domains (``Mean'' column).
We discuss our findings.

\noindent\textbf{Different pretrainings.}
For the dimensionality of 768-D produced by the ViT-B backbone, DINOv2 is the best performing pretraining method compared to CLIP and IN.
We can partially attribute this to DINOv2 having also been trained on parts of our training set (see~\cite{oquab2023dinov2} for details of DINOv2 pretraining data).
MultiGrain, that employs a CNN ResNet50 backbone, underperforms the others, despite the much higher dimensionality.
PCA-Whitening further harms performance; in the Appendix we also compare it with a random linear projection to 64-D, showing that it performs on par with PCA-Whitening.
Overall, even the much higher dimensional off-the-shelf embeddings underperform our finetuned 64-D embeddings.

\noindent\textbf{Oracle Specialist embedding.}
In this setting, 8 models were trained, one for each domain.
The domain of the query image at test time is used, so the corresponding specialist of that domain is used to extract its embedding, as well as the embeddings for the entire merged index set.
The best average performance is achieved, but it only consitutes an unrealistic setting, since Oracle is used to select the correct specialist.

\begin{figure*}[ht]
\begin{center}
\begin{tabular}{cc}
 \scalebox{0.61}{
    \begin{tabular}{llllll}
    \hspace{17pt} Query \hspace{31pt} 1st rank \hspace{23pt} 2nd rank \hspace{23pt} 3rd rank \hspace{23pt} 4th rank \hspace{23pt} 5th rank \\
    \multicolumn{6}{l}{\includegraphics[scale = 0.29]{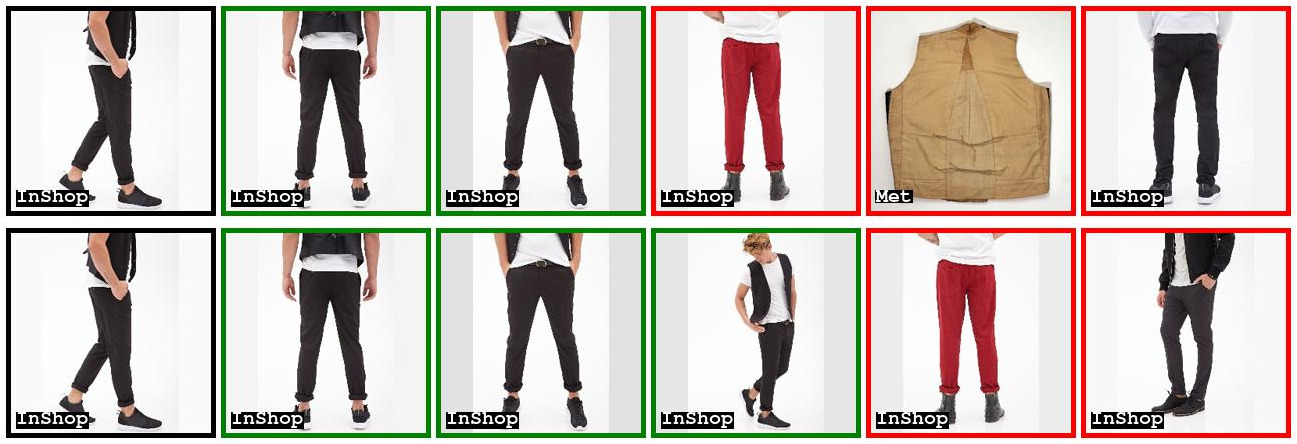}}\\
    \\
    \multicolumn{6}{l}{\includegraphics[scale = 0.29]{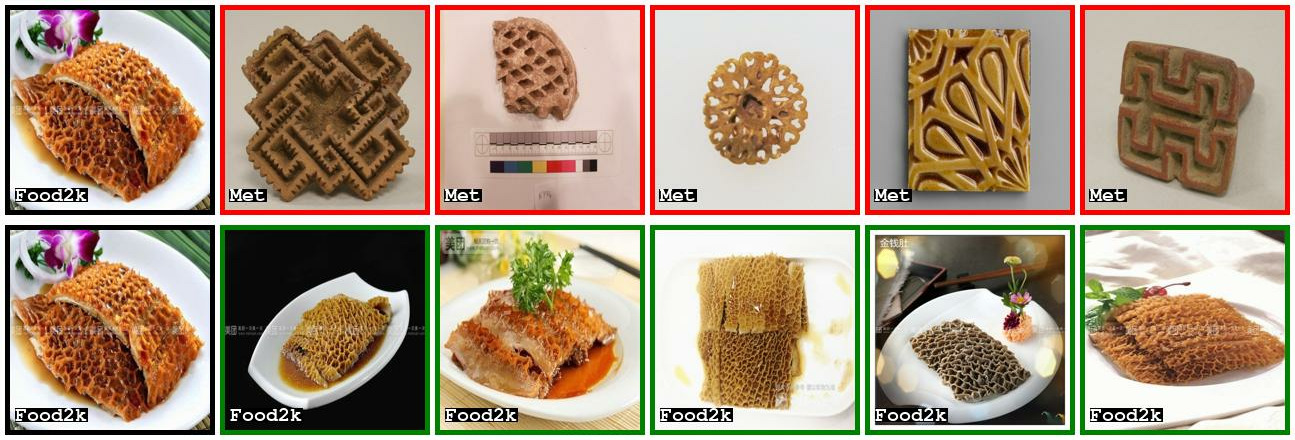}}\\
    \\
    \multicolumn{6}{l}{\includegraphics[scale = 0.29]{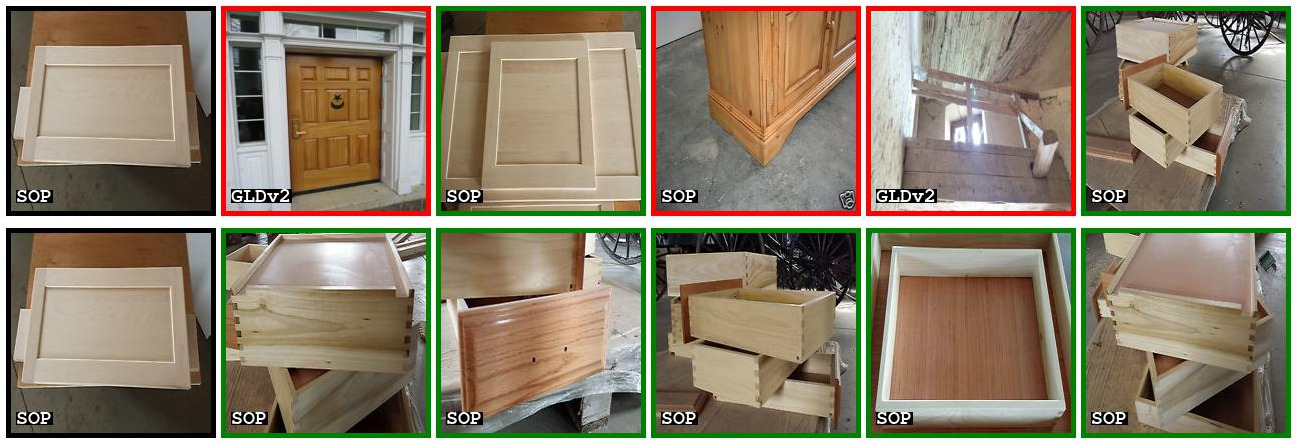}}\\
    \\
    \end{tabular}
    }
& 
 \scalebox{0.61}{
    \begin{tabular}{llllll}
    \hspace{17pt} Query \hspace{31pt} 1st rank \hspace{23pt} 2nd rank \hspace{23pt} 3rd rank \hspace{23pt} 4th rank \hspace{23pt} 5th rank \\
    \multicolumn{6}{l}{\includegraphics[scale = 0.29]{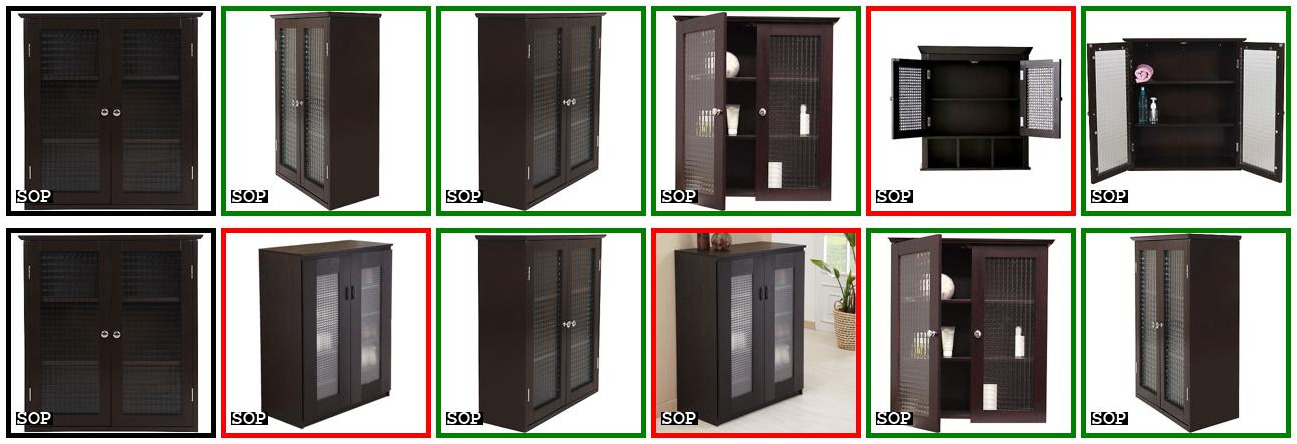}}\\
    \\
    \multicolumn{6}{l}{\includegraphics[scale = 0.29]{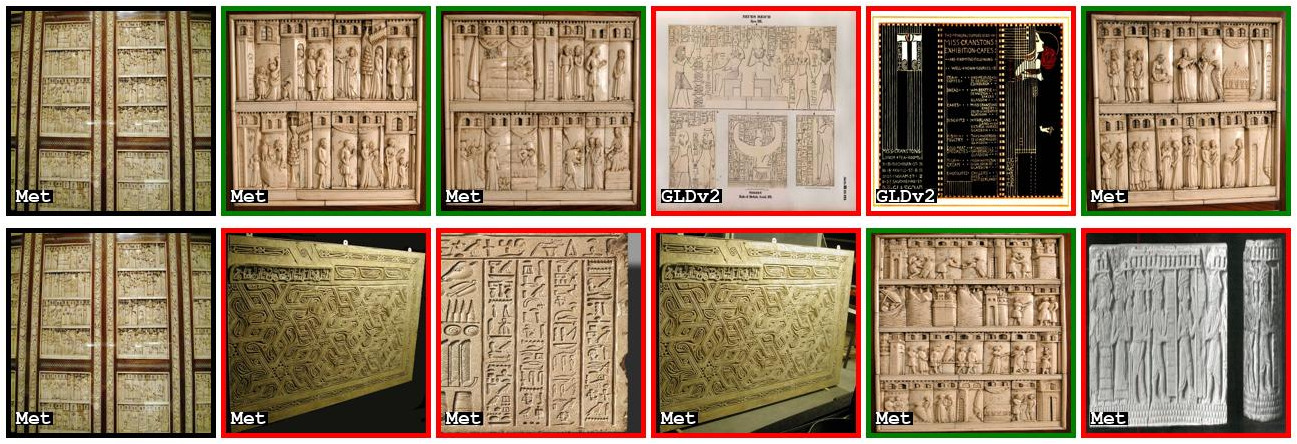}}\\
    \\
    \multicolumn{6}{l}{\includegraphics[scale = 0.29]{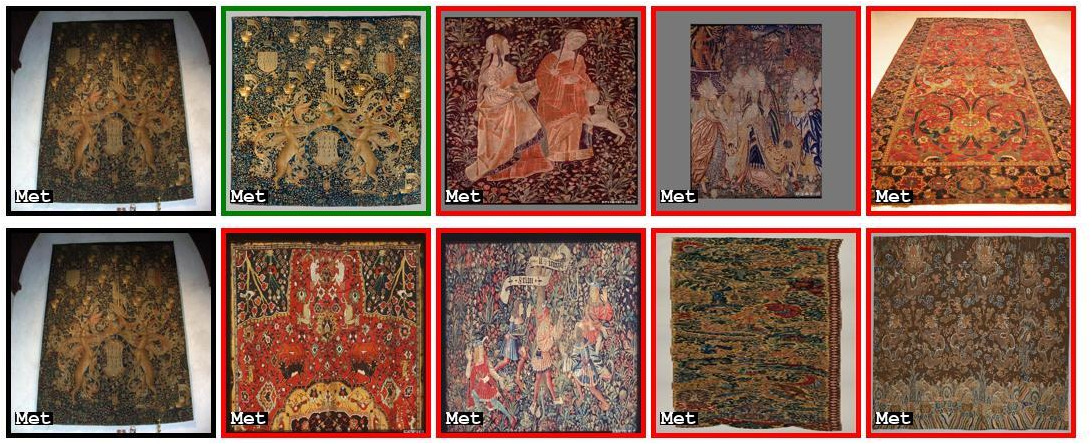}}\\
    \\
    \end{tabular}
    }
\end{tabular}
\vspace{-20pt}
\caption{
\textbf{Retrieval results for example queries.} 
Each column contains 3 queries, for each query the top row corresponds to nearest neighbors from the Specialist CLIP+Oracle embedding, bottom row from the universal model CLIP+USCSS.
A correct neighbor (coming from the same class) is indicated by green border; otherwise by red.
The number of neighbors shown next to each query equals the number that is taken into account for the computation of the mMP@5 metric for that query, \ie $\min (n_q, 5)$, where $n_q$ is the number of index images from the same class as the query.
The domain that the image comes from is shown on each image.
On the left column, examples of cross-domain failure for the Specialist CLIP+Oracle model are shown; on the right examples of failure for the universal model are shown, indicating degraded fine-grained capabilities.
\vspace{-5pt}
\label{fig:retrieval_main}
}
\end{center}
\end{figure*}
\noindent\textbf{Universal embedding models.}
The baseline techniques of universal embedding training examined in this work are direct extensions of specialist training methods. 
During training no expert knowledge was exploited (\eg the Met dataset would clearly benefit from strong geometric augmentations).
Still, the final performance is close to the Specialists performance, or even surpasses the Specialists in the SOP and InShop domains. 
This is remarkable given that 8 times less parameters are used compared to the corresponding Specialist Oracle
, and also that no knowledge of the test time domain is utilized.

We observe the following: (i) The universal models reach the same performance of validation retrieval metrics (R@1 and mMP@5) on most domains faster than the corresponding specialist (in terms of total optimization steps performed for training samples of that domain); we attribute this to the sharing of useful features across domains, 
(ii) Different domains overfit at different rates, as observed on the validation retrieval metrics; this has also been reported in~\cite{feng2020unifying},
(iii) Different domains benefit differently from the pretraining method, \ie CARS196 benefits a lot more from CLIP pretraining, while iNat domain benefits more from IN pretraining.
(iv) For a given sampling scheme (RR), the different classifiers (Joint - J and Separate - S) produce different results across domains.
For example, the SOP and InShop domains benefit the most when trained using the joint classifier. 
On the other hand, the Met domain benefits by the use of a separate classifier; however this specific domain suffers from a very low performance compared to the specialist.
Preliminary experiments on pairs of datasets revealed that the combination of GLDv2 and Met makes the training difficult for the Met domain.
(v) Regarding sampling strategies, sampling based on dataset size performs the worst on average, while the RR methods that balance the domains improve average performance consistently.
Sampling according to the number of steps that the specialist maximizes its performance (SS) performs the best on average, however it produces the highest standard deviation across seeds, as the number of steps each specialist reaches maximum performance at is also dependent on the seed itself.

Qualitative examples for the comparison of retrieval between the CLIP+USCSS model and the Specialist CLIP+Oracle are shown in Figure~\ref{fig:retrieval_main}.
We observe the cross-domain failure for Specialist CLIP+Oracle model, as shown in some examples in the left column.
This can be attributed to the universal model seeing all of the domains at train time, while the specialist model fails to handle images that are out of its train domain distribution.
Also, the universal model shows degraded instance-level/fine-grained discrimination capabilities, as shown in the examples in the right column.
This can be attributed to the universal model having to utilize the same capacity to learn all the domains that a specialist utilizes for one.
\section{The Universal Embedding Challenge}
\label{sec:challenge}
Motivated by the strong need of having a single universal embedding for various industrial applications, complementary to the proposed Universal Embedding Dataset, we conducted the Google Universal Image Embedding Challenge in Kaggle.
This competition stimulated research interests in developing ideas and methods in training universal image representations, which we introduce in detail in this section.
\begin{table*}[ht]
  \centering
  \scalebox{0.68}{
\begin{tabular}{cccc}
\toprule
Model & Team & Techniques & mMP@5 (private split) \\
\midrule
Top 1 & cuilab.ai \cite{challengetop1} & CLIP ViT-H/14 $\to$ Multi-domain data + multi-stage finetuning & 72.8 \\
Top 2 & Xiao \cite{challengetop2} & CLIP ViT-H/14 $\to$ Multi-domain data + stratified learning rates & 70.9 \\
Top 3 & - - - \cite{challengetop3} & CLIP ViT-H/14 $\to$ Multi-domain data + dropout rate ensembling & 69.2 \\
Top 4 & Ivan \& Simjeg \& CLIP-Art \cite{challengetop4} & CLIP ViT-H/14 + CLIP ViT-L/14 $\to$ Multi-domain data + model size ensembling & 68.8 \\
Top 5 & NS embedding \cite{challengetop5} & CLIP ViT-H/14 $\to$ Multi-domain data + adding image heuristics (height, width) & 68.8 \\
Top 6 & IRonCLIP \cite{challengetop6} & CLIP ViT-H/14 $\to$ Multi-domain data + multi-stage finetuning + Test time augmentation (TTA) & 68.5  \\
CLIP ViT-H/14 & - & Reference model \alert{\textbf{(1024-D)}} & 62.1  \\
\bottomrule
\end{tabular}
}
  \vspace{-7pt}
  \caption{Summary of techniques used in the top 6 winning solutions in the Google Universal Image Embedding Challenge and their mMP@5 scores on the private split in the challenge. We also include the score of the pre-trained model (CLIP ViT-H/14) in the last row, as a reference.
  \label{tab:challenge_techniques}
  \vspace{-15pt}
  }
\end{table*}
\noindent\textbf{Challenge dataset.}
For the evaluation dataset for the challenge, instead of using the one presented in this paper, we introduced a separate benchmark composed of a query set with $5k$ images and an index set with $200k$ images.
This dataset covered $11$ image domains, including: apparel \& accessories, packaged goods, landmarks, furniture \& home decor, storefronts, dishes, artwork, toys, memes, illustrations and cars -- which are considered of significant importance for industrial applications.
These query and index images were collected and annotated with fine-grained and instance-level labels by human annotators.
The distribution of the domains of query images was disclosed to participants in the challenge, who could tune their methods with this information.
The queries were split into two subsets: $30\%$ were used for scoring models while the competition was running, which gave feedback to participants on the quality of their submissions.
The rest $70\%$ queries were kept separate and only used for the final scoring, once the competition closed for new submissions.
Given that this dataset is collected for industrial applications, it is not possible to release the raw images to participants. 

In terms of training data, a crucial obstacle in industrial applications is on how to select the most relevant images for training.
To create a similar setup, in this challenge, we didn't provide any specific guidance on the training datasets to use: selecting the right datasets was one of the challenges that participants had to face.

We would like to highlight important differences between the public dataset proposed in this paper and the dataset used in the challenge.
Firstly, in the challenge, we avoid using any publicly available datasets to prevent overfitting. Secondly, the challenge dataset is collected for industrial applications, and one of the goals is to verify that techniques that work on this dataset can also be applicable to the public datasets proposed in this paper. Thirdly, in many industrial applications, the training dataset is ambiguous and only the distribution of evaluation is known. This challenge is to mimic this setting to encourage novel ideas under these scenarios.

\noindent\textbf{Challenge setup.}
Similar to the setup discussed in this paper, the challenge focused on image retrieval task using 64-D image features, with the mMP@5 metric defined in Equation \ref{eq:3}. The model evaluations were conducted through a scoring notebook hosted on our servers which ran on GPU P100 chips with 16G memory.
During submission, participants were asked to upload their models trained with either PyTorch \cite{paszke2019} or Tensorflow \cite{martin2016}.
Based on the uploaded model, the scoring notebook would run feature extraction and metric computation on the evaluation dataset.
We also set a loose runtime limit of 9 hours for the scoring notebook to provide flexibility of the model size and use this as an incentive for researchers to explore different model architectures.
 
\begin{table*}[ht]
  \centering
  \scalebox{0.632}{
\begin{tabular}{c|cc|cc|cc|cc|cc|cc|cc|cc|cc}
\hline
  & \multicolumn{2}{c|}{Food2k}& \multicolumn{2}{c|}{CARS196}& \multicolumn{2}{c|}{SOP}&\multicolumn{2}{c|}{InShop}&\multicolumn{2}{c|}{iNat}& \multicolumn{2}{c|}{Met}& \multicolumn{2}{c|}{GLDv2}& \multicolumn{2}{c|}{Rp2k}&  \multicolumn{2}{c}{Mean}\\ 
\hline
Model &mMP@5 & R@1 & mMP@5 & R@1 & mMP@5 & R@1 & mMP@5 & R@1 & mMP@5 & R@1 & mMP@5 & R@1 & mMP@5 & R@1 & mMP@5 & R@1 & mMP@5 & R@1\\
\hline
Top 1& 41.2 & 53.0 & 96.2 & 97.3 & 54.7 & 70.7 & 52.7 & 78.5 & \textbf{68.7} & \textbf{74.9} & \textbf{45.0} & \textbf{55.1} & \textbf{42.3} & \textbf{53.7} & 83.3 & 91.4 & 60.5 & 71.8 \\

Top 2& \textbf{46.4} & \textbf{57.3} & 97.1 & 97.7 & \textbf{65.1} & \textbf{79.9} & \textbf{66.4} & \textbf{87.1} & 61.2 & 68.2 & 39.7 & 48.8 & 37.6 & 49.5 & \textbf{87.1} & \textbf{93.4} & \textbf{62.6} & \textbf{72.7} \\
Top 3& 38.2 & 49.9 & 93.7 & 95.2 & 53.7 & 70.0 & 50.2 & 75.5 & 61.5 & 68.1 & 18.6 & 23.8 & 39.8 & 52.4 & 76.7 & 87.8 & 54.0 & 65.3 \\
Top 4& 37.3 & 48.8 & 92.2 & 95.6 & 49.6 & 67.1 & 46.5 & 72.6 & 63.0 & 69.5 & 40.1 & 51.4 & 35.3 & 48.4 & 78.9 & 89.4 & 55.4 & 67.9 \\
Top 5& 41.5 & 54.3 & 96.1 & 97.2 & 54.9 & 71.7  & 51.1 & 76.5 & 67.5 & 73.8 & 45.5 & 56.2 & 34.7 & 47.4 & 73.4 & 87.1 & 58.1 & 70.5 \\
Top 6& 38.8 & 49.8 & \textbf{97.5} & \textbf{97.8} & 58.8 & 74.7 & 60.2 & 82.8 & 57.3 & 64.4 & 25.9 & 31.5 & 35.4 & 46.4 & 83.3 & 91.7 & 57.1 & 67.4 \\
\hline
\end{tabular}
}
  \vspace{-7pt}
  \caption{
  Retrieval metrics on the test set of the Universal Embedding Dataset for the top 6 solutions of the Google Universal Image Embedding Challenge.
  Note that in obtaining these metrics, we followed the same steps from the challenge submissions to preprocess the test images, which are not necessarily the same as the ones used for the models in this paper.
  \label{tab:challenge_eval}
  \vspace{-5pt}
  }
\end{table*}
\subsection{Challenge results and findings}

We ran the Universal Embedding Challenge for three months. At the end, this challenge attracted around $1k$ teams with in total $21k$ model submissions. We summarize the techniques used in the top 6 solutions in Table \ref{tab:challenge_techniques}. All teams used a pre-trained CLIP model \cite{radford2021learning} as the backbone. To meet the feature dimension requirement of the challenge, shallow projection layers were added on top of the pre-trained model to produce $64$-D embeddings. The models were then finetuned on multi-domain datasets using standard augmentation techniques and supervised by classification tasks with ArcFace loss \cite{deng2019arcface}. This training recipe as a result showed better performance than directly applying PCA to the output features from the pre-trained model, as presented by a detailed study from the Top 4 team \cite{challengetop4}. In the following, we highlight several key findings and discuss them in detail.

\noindent\textbf{Improved pre-training via image-text foundation models.} All the winning models first initialized from and then finetuned the image backbone
(ViT-H/14 or ViT-L/14 \cite{dosovitskiy2020image}) of the CLIP model pre-trained on the LAION2B dataset \cite{schuhmann2022laionb}.
These were some of the first explorations leveraging image-text foundation models as a central building block for complex image retrieval models (previous work \cite{wu2022,ibrahimi2022} had started investigating this direction). The success of adopting such foundation models here indicates that rich detailed information contained in text can be beneficial to vision-only models that need to be sensitive to fine-grained details.
In contrast, most previous embedding models had been pre-trained on datasets with coarse categories, such as ImageNet \cite{deng2009}.
Moreover, these foundation models are usually trained with large-scale datasets which can naturally contain many domains -- this is another reason that makes these models suitable for generic representation learning. Our exploration of CLIP models presented in Section~\ref{sec:benchmarking} is largely motivated by this observation.

\noindent\textbf{Improved fine-tuning.}
In Table \ref{tab:challenge_techniques}, we present the evaluations of the finetuned models provided by the top teams and the CLIP ViT-H/14 model in the last row, as a reference. By comparing the mMP@5 scores, we show that this pre-trained model, though having a larger feature dimension of 1024, obtains inferior performance by up to $10.7\%$, which demonstrates the effectiveness of finetuning.
For the finetuning techniques used in the top solutions, we found that treating the backbone and the shallow projection and classifier layers separately, by either making finetuning multi-stage or using different learning rates, is very necessary. For instance in the Top 2 solution \cite{challengetop2}, the learning rates for the $64$-D projection layer and the classifiers are set to be $10^3$ times larger than that of the backbone, and with this stratified learning rate setup, the final mMP@5 score is significantly increased to $70.9$ compared to $62.1$ of the pre-trained model. This training strategy also makes intuitive sense. Firstly, the extra projection layer and classifier layers are randomly initialized during finetuning, thus requiring either higher learning rate or a ``warm up'' stage in order to properly train the weights. Secondly, given that the initial backbone weights are already well-trained, using large learning rates might cause undesired overfitting and destroy its generic knowledge. This aligns with recent trend that using frozen or slightly finetuned \cite{minderer2022, ghiasi2022, almazan2022granularity} pre-trained models can improve the performance for different tasks.
Our experiments presented in this paper also adopt the same strategy. Furthermore, model soup \cite{wortsman22a} and model ensembling 
are also experimented by several teams. In particular, ensembling features trained with various dropout rates or with different backbone sizes proved to help improve the models' performance.

\noindent\textbf{Training set selection.}
To properly finetune the pre-trained model towards the challenge's evaluation, selecting the right set of training data is crucial. We notice that participants explored a variety of datasets in different ways, and the procedure for data selection is designed to match the distribution of the query set. Given multiple datasets containing images from different domains, we observe the Top 1 team \cite{challengetop1} used greedy search algorithm for datasets and only keep the ones that help the performance. The Top 2 team \cite{challengetop2} conducted very detailed investigation on data balancing and mixing, and they found that filtering out rare classes and capping the maximum number of images per class were helpful. 
The Top 6 team \cite{challengetop6} leveraged the pre-trained CLIP model to generate labels for noisy datasets.

\subsection{Top challenge solutions on our benchmark}
To better understand the performance of the challenge winning solutions on the benchmarks proposed in this paper, in Table \ref{tab:challenge_eval} we present the evaluation of the top 6 solutions on the proposed test sets.
These models on average outperform the baseline presented in this paper since they are 1) much larger in size, 2) initialized from a model pre-trained on larger-scale datasets, and 3) trained with more involved finetuning procedures. 
We note that the second place solution in fact achieves the highest average performance in our benchmarks. 
In domains such as CARS196, Met, GLDv2 and Rp2k, the challenge solutions outperform our specialist models and the universal models. However, in domains such as Food2k, SOP, and Inshop, our trained models perform better. This is because these domains are not the majority of the challenge evaluation dataset and therefore the submitted models were either not explicitly trained on the related datasets or not optimized toward these domains. By comparing the numbers in Tables \ref{tab:challenge_techniques} and \ref{tab:challenge_eval}, we also conclude that the evaluation between the challenge dataset and the dataset proprosed in this paper are correlated given that the top 2 winners in the challenge also have the highest metrics in our benchmarks.
\section{Conclusions}
\label{sec:conclusions}

In this work, the novel large-scale UnED dataset for training and evaluating Universal Embedding models was introduced in order to stimulate research in the area.
A comprehensive benchmarking was performed and reference models for future comparison were implemented and discussed. 
The universal training baselines introduced
revealed some of the challenges as well as the benefits of learning the representation for multiple domains simultaneously.
Finally, a public challenge on universal embeddings was conducted, techniques exploited by the top ranked teams were discussed and the winning methods were evaluated on the proposed benchmark.
The proposed splits will be released on the project website, while the implemented baseline methods, as well as the evaluation scripts, will be released under two frameworks (Scenic~\cite{dga+21} and PyTorch~\cite{paszke2019}).
We expect that the metric learning field will significantly progress by focusing on learning universal image representations.

\newpage
{\small
\bibliographystyle{ieee_fullname}
\bibliography{egbib}

\begin{thebibliography}{10}\itemsep=-1pt

\bibitem{aliproductschallenge}
{AliProducts Challenge: Large-scale Product Recognition}.
\newblock {\em
  \url{https://tianchi.aliyun.com/competition/entrance/231780/introduction?lang=en-us}},
  2020.

\bibitem{deepfashion2challenge}
{DeepFashion2 Challenge}.
\newblock {\em
  \url{https://sites.google.com/corp/view/cvcreative2020/deepfashion2}}, 2020.

\bibitem{ebaychallenge}
{eBay eProduct Visual Search Challenge}.
\newblock {\em
  \url{https://eval.ai/web/challenges/challenge-page/888/leaderboard/2352}},
  2021.

\bibitem{gld21challenge}
{Google Landmark Retrieval Challenge 2021}.
\newblock {\em \url{https://www.kaggle.com/c/landmark-retrieval-2021}}, 2021.

\bibitem{products}
Amazon app camera search.
\newblock {\em Available online:
  \url{https://www.amazon.com/b?ie=UTF8&node=17387598011}}, 2023.

\bibitem{art}
Art recognition.
\newblock {\em Available online: \url{https://art-recognition.com/}}, 2023.

\bibitem{bixby}
Bixby vision.
\newblock {\em Available online:
  \url{https://www.samsung.com/global/galaxy/apps/bixby/vision/}}, 2023.

\bibitem{cars}
Car recognition.
\newblock {\em Available online:
  \url{https://play.google.com/store/apps/details?id=com.asdev.carmake}}, 2023.

\bibitem{googlelens}
{Google Lens}.
\newblock {\em Available online: \url{https://lens.google/}}, 2023.

\bibitem{landmarks}
Landmark detection.
\newblock {\em Available online:
  \url{https://cloud.google.com/vision/docs/detecting-landmarks}}, 2023.

\bibitem{natural}
Picturethis - plant identifier.
\newblock {\em Available online:
  \url{https://apps.apple.com/us/app/picturethis-plant-identifier/id1252497129}},
  2023.

\bibitem{food}
Snapcalorie.
\newblock {\em Available online: \url{https://www.snapcalorie.com/}}, 2023.

\bibitem{martin2016}
Martin Abadi, Paul Barham, Jianmin Chen, Zhifeng Chen, Andy Davis, Jeffrey
  Dean, Matthieu Devin, Sanjay Ghemawat, Geoffrey Irving, Michael Isard,
  Manjunath Kudlur, Josh Levenberg, Rajat Monga, Sherry Moore, Derek~G. Murray,
  Benoit Steiner, Paul Tucker, Vijay Vasudevan, Pete Warden, Martin Wicke, Yuan
  Yu, and Xiaoqiang Zheng.
\newblock {TensorFlow: A system for large-scale machine learning}.
\newblock In {\em 12th USENIX Symposium on Operating Systems Design and
  Implementation (OSDI 16)}, pages 265--283, 2016.

\bibitem{almazan2022granularity}
J. Almazan, B. Ko, G. Gu, D. Larlus, and Y. Kalantidis.
\newblock {Granularity-aware Adaptation for Image Retrieval over Multiple
  Tasks}.
\newblock In {\em Proc. ECCV}, 2022.

\bibitem{challengetop3}
N. Aoki and Y. Namba.
\newblock {3rd Place Solution for Google Universal Image Embedding}.
\newblock {\em arXiv:2210.09296}, 2022.

\bibitem{berman2019multigrain}
M. Berman, H. J{\'e}gou, A. Vedaldi, I. Kokkinos, and M. Douze.
\newblock {MultiGrain: a Unified Image Embedding for Classes and Instances}.
\newblock {\em arXiv:1902.05509}, 2019.

\bibitem{brown2020smooth}
Andrew Brown, Weidi Xie, Vicky Kalogeiton, and Andrew Zisserman.
\newblock Smooth-ap: Smoothing the path towards large-scale image retrieval.
\newblock In {\em Computer Vision--ECCV 2020: 16th European Conference,
  Glasgow, UK, August 23--28, 2020, Proceedings, Part IX 16}, pages 677--694.
  Springer, 2020.

\bibitem{cao2020unifying}
B. Cao, A. Araujo, and J. Sim.
\newblock {Unifying Deep Local and Global Features for Image Search}.
\newblock In {\em Proc. ECCV}, 2020.

\bibitem{chen2022deep}
W. Chen, Liu Y, W. Wang, E.~M. Bakker, T. Georgiou, P. Fieguth, L. Liu, and
  M.~S. Lew.
\newblock {Deep Learning for Instance Retrieval: A Survey}.
\newblock {\em TPAMI}, 2022.

\bibitem{challengetop4}
M. Conde, I. Aerlic, and S. Jégou.
\newblock {General Image Descriptors for Open World Image Retrieval using ViT
  CLIP}.
\newblock {\em arXiv:2210.11141}, 2022.

\bibitem{dga+21}
Mostafa Dehghani, Alexey Gritsenko, Anurag Arnab, Matthias Minderer, and Yi
  Tay.
\newblock Scenic: A jax library for computer vision research and beyond.
\newblock In {\em Proceedings of the IEEE/CVF Conference on Computer Vision and
  Pattern Recognition (CVPR)}, pages 21393--21398, 2022.

\bibitem{deng2009}
Jia Deng, Wei Dong, Richard Socher, Li-Jia Li, Kai Li, and Li Fei-Fei.
\newblock {ImageNet: A large-scale hierarchical image database}.
\newblock In {\em Proc. CVPR}, pages 248--255, 2009.

\bibitem{deng2019arcface}
J. Deng, J. Guo, N. Xue, and S. Zafeiriou.
\newblock {ArcFace: Additive Angular Margin Loss for Deep Face Recognition}.
\newblock In {\em Proc. CVPR}, 2019.

\bibitem{dosovitskiy2020image}
Alexey Dosovitskiy, Lucas Beyer, Alexander Kolesnikov, Dirk Weissenborn,
  Xiaohua Zhai, Thomas Unterthiner, Mostafa Dehghani, Matthias Minderer, Georg
  Heigold, Sylvain Gelly, et~al.
\newblock An image is worth 16x16 words: Transformers for image recognition at
  scale.
\newblock {\em arXiv preprint arXiv:2010.11929}, 2020.

\bibitem{feng2020unifying}
Yang Feng, Futang Peng, Xu Zhang, Wei Zhu, Shanfeng Zhang, Howard Zhou, Zhen
  Li, Tom Duerig, Shih-Fu Chang, and Jiebo Luo.
\newblock {Unifying Specialist Image Embedding into Universal Image Embedding}.
\newblock {\em arXiv:2003.03701}, 2020.

\bibitem{ghiasi2022}
G. Ghiasi, X. Gu, Y. Cui, and T.-Y. Lin.
\newblock {Scaling Open-Vocabulary Image Segmentation With Image-Level Labels}.
\newblock In {\em Proc. ECCV}, 2021.

\bibitem{challengetop6}
S. Gkelios, A. Kastellos, and S. Chatzichristofis.
\newblock {6th Place Solution to Google Universal Image Embedding}.
\newblock {\em arXiv:2210.09377}, 2022.

\bibitem{gordo2016deep}
A. Gordo, J. Almazan, J. Revaud, and D. Larlus.
\newblock {Deep Image Retrieval: Learning Global Representations for Image
  Search}.
\newblock In {\em Proc. ECCV}, 2016.

\bibitem{hu2015deep}
J. Hu, J. Lu, and Y.-P. Tan.
\newblock {Deep Transfer Metric Learning}.
\newblock In {\em Proc. CVPR}, 2015.

\bibitem{challengetop2}
X. Huang and Q. Li.
\newblock {2nd Place Solution to Google Universal Image Embedding}.
\newblock {\em arXiv:2210.08735}, 2022.

\bibitem{ibrahimi2022}
S. Ibrahimi, A. Sors, R.~Sampaio de Rezende, and S. Clinchant.
\newblock {Learning with Label Noise for Image Retrieval by Selecting
  Interactions}.
\newblock In {\em Proc. WACV}, 2022.

\bibitem{jegou2012negative}
H. J{\'e}gou and O. Chum.
\newblock {Negative Evidences and Co-Occurences in Image Retrieval: The Benefit
  of PCA and Whitening}.
\newblock In {\em Proc. ECCV}, 2012.

\bibitem{kemelmacher2016megaface}
I. Kemelmacher-Shlizerman, S. Seitz, D. Miller, and E. Brossard.
\newblock {The MegaFace Benchmark: 1 Million Faces for Recognition at Scale}.
\newblock In {\em Proc. CVPR}, 2016.

\bibitem{kim2022adaface}
M. Kim, A.~K. Jain, and X. Liu.
\newblock {AdaFace: Quality Adaptive Margin for Face Recognition}.
\newblock In {\em Proc. CVPR}, 2022.

\bibitem{krause20133d}
J. Krause, M. Stark, J. Deng, and L. Fei-Fei.
\newblock {3D Object Representations for Fine-Grained Categorization}.
\newblock In {\em Proc. ICCV Workshops}, 2013.

\bibitem{lee2022correlation}
S. Lee, H. Seong, S. Lee, and E. Kim.
\newblock {Correlation Verification for Image Retrieval}.
\newblock In {\em Proc. CVPR}, 2022.

\bibitem{levi2021rethinking}
E. Levi, T. Xiao, X. Wang, and T. Darrell.
\newblock {Rethinking Preventing Class-Collapsing in Metric Learning with
  Margin-based Losses}.
\newblock In {\em Proc. ICCV}, 2021.

\bibitem{li2014deepreid}
W. Li, R. Zhao, T. Xiao, and X. Wang.
\newblock {DeepReID: Deep Filter Pairing Neural Network for Person
  Re-Identification}.
\newblock In {\em Proc. CVPR}, 2014.

\bibitem{lac+21}
Valerii Likhosherstov, Anurag Arnab, Krzysztof Choromanski, Mario Lucic, Yi
  Tay, Adrian Weller, and Mostafa Dehghani.
\newblock Polyvit: Co-training vision transformers on images, videos and audio.
\newblock {\em arXiv preprint arXiv:2111.12993}, 2021.

\bibitem{liu2016deepfashion}
Ziwei Liu, Ping Luo, Shi Qiu, Xiaogang Wang, and Xiaoou Tang.
\newblock Deepfashion: Powering robust clothes recognition and retrieval with
  rich annotations.
\newblock In {\em Proceedings of IEEE Conference on Computer Vision and Pattern
  Recognition (CVPR)}, 2016.

\bibitem{maze2018ijbc}
B. Maze, J. Adams, J.~A. Duncan, N. Kalka, T. Miller, C. Otto, A.~K. Jain,
  W.~T. Niggel, J. Anderson, and Jordan Cheney.
\newblock {Iarpa Janus Benchmark–C: Face Dataset and Protocol}.
\newblock {\em ICB}, 2018.

\bibitem{min2023large}
Weiqing Min, Zhiling Wang, Yuxin Liu, Mengjiang Luo, Liping Kang, Xiaoming Wei,
  Xiaolin Wei, and Shuqiang Jiang.
\newblock Large scale visual food recognition.
\newblock {\em IEEE Transactions on Pattern Analysis and Machine Intelligence},
  2023.

\bibitem{minderer2022}
M. Minderer, A. Gritsenko, A. Stone, M. Neumann, D. Weissenborn, A.
  Dosovitskiy, A. Mahendran, A. Arnab, M. Dehghani, Z. Shen, X. Wang, X. Zhai,
  T. Kipf, and N. Houlsby.
\newblock {Simple Open-Vocabulary Object Detection}.
\newblock In {\em Proc. ECCV}, 2022.

\bibitem{musgrave2020reality}
K. Musgrave, S. Belongie, and Ser-Nam Lim.
\newblock {A Metric Learning Reality Check}.
\newblock In {\em Proc. ECCV}, 2020.

\bibitem{oquab2023dinov2}
Maxime Oquab, Timoth{\'e}e Darcet, Th{\'e}o Moutakanni, Huy Vo, Marc
  Szafraniec, Vasil Khalidov, Pierre Fernandez, Daniel Haziza, Francisco Massa,
  Alaaeldin El-Nouby, et~al.
\newblock Dinov2: Learning robust visual features without supervision.
\newblock {\em arXiv preprint arXiv:2304.07193}, 2023.

\bibitem{challengetop5}
N. Ota, S. Yokoi, and S. Yamaoka.
\newblock {5th Place Solution to Kaggle Google Universal Image Embedding
  Competition}.
\newblock {\em arXiv:2210.09495}, 2022.

\bibitem{paszke2019}
Adam Paszke, Sam Gross, Francisco Massa, Adam Lerer, James Bradbury, Gregory
  Chanan, Trevor Killeen, Zeming Lin, Natalia Gimelshein, Luca Antiga, Alban
  Desmaison, Andreas K\"{o}pf, Edward Yang, Zach DeVito, Martin Raison, Alykhan
  Tejani, Sasank Chilamkurthy, Benoit Steiner, Lu Fang, Junjie Bai, and Soumith
  Chintala.
\newblock {\em {PyTorch: An Imperative Style, High-Performance Deep Learning
  Library}}.
\newblock 2019.

\bibitem{peng2020rp2k}
Jingtian Peng, Chang Xiao, and Yifan Li.
\newblock Rp2k: A large-scale retail product dataset forfine-grained image
  classification.
\newblock {\em arXiv preprint arXiv:2006.12634}, 2020.

\bibitem{qian2015fine}
Q. Qian, R. Jin, S. Zhu, and Y. Lin.
\newblock {Fine-Grained Visual Categorization via Multi-stage Metric Learning}.
\newblock In {\em Proc. CVPR}, 2015.

\bibitem{radenovic2018revisiting}
F. Radenovi{\'c}, A. Iscen, G. Tolias, Y. Avrithis, and O. Chum.
\newblock {Revisiting Oxford and Paris: Large-Scale Image Retrieval
  Benchmarking}.
\newblock In {\em Proc. CVPR}, 2018.

\bibitem{radford2021learning}
Alec Radford, Jong~Wook Kim, Chris Hallacy, Aditya Ramesh, Gabriel Goh,
  Sandhini Agarwal, Girish Sastry, Amanda Askell, Pamela Mishkin, Jack Clark,
  et~al.
\newblock Learning transferable visual models from natural language
  supervision.
\newblock In {\em International conference on machine learning}, pages
  8748--8763. PMLR, 2021.

\bibitem{roth2022integrating}
K. Roth, O. Vinyals, and Z. Akata.
\newblock {Integrating Language Guidance into Vision-based Deep Metric
  Learning}.
\newblock In {\em Proc. CVPR}, 2022.

\bibitem{schall2021gpr1200}
K. Schall, K.~U. Barthel, N. Hezel, and K. Jung.
\newblock {GPR1200: A Benchmark for General-Purpose Content-Based Image
  Retrieval}.
\newblock In {\em Proc. International Conference on Multimedia Modeling}, 2022.

\bibitem{schroff2015facenet}
F. Schroff, D. Kalenichenko, and J. Philbin.
\newblock { FaceNet: A Unified Embedding for Face Recognition and Clustering}.
\newblock In {\em Proc. CVPR}, 2015.

\bibitem{schuhmann2022laionb}
Christoph Schuhmann, Romain Beaumont, Richard Vencu, Cade~W Gordon, Ross
  Wightman, Mehdi Cherti, Theo Coombes, Aarush Katta, Clayton Mullis, Mitchell
  Wortsman, Patrick Schramowski, Srivatsa~R Kundurthy, Katherine Crowson,
  Ludwig Schmidt, Robert Kaczmarczyk, and Jenia Jitsev.
\newblock {LAION-5B: An open large-scale dataset for training next generation
  image-text models}.
\newblock In {\em Proc. NeurIPS}, 2022.

\bibitem{seidenschwarz2021learning}
J. Seidenschwarz, I. Elezi, and L. Leal-Taixe.
\newblock {Learning Intra-Batch Connections for Deep Metric Learning}.
\newblock In {\em Proc. ICML}, 2021.

\bibitem{challengetop1}
S. Shao and Q. Cui.
\newblock {1st Place Solution in Google Universal Images Embedding}.
\newblock {\em arXiv:2210.08473}, 2022.

\bibitem{song2016deep}
Hyun~Oh Song, Yu Xiang, Stefanie Jegelka, and Silvio Savarese.
\newblock {Deep Metric Learning via Lifted Structured Feature Embedding}.
\newblock In {\em Proc. CVPR}, 2016.

\bibitem{steiner2021train}
Andreas Steiner, Alexander Kolesnikov, Xiaohua Zhai, Ross Wightman, Jakob
  Uszkoreit, and Lucas Beyer.
\newblock How to train your vit? data, augmentation, and regularization in
  vision transformers.
\newblock {\em arXiv preprint arXiv:2106.10270}, 2021.

\bibitem{tan2021instance}
F. Tan, J. Yuan, and V. Ordonez.
\newblock {Instance-level Image Retrieval using Reranking Transformers}.
\newblock In {\em Proc. ICCV}, 2021.

\bibitem{van2018inaturalist}
Grant Van~Horn, Oisin Mac~Aodha, Yang Song, Yin Cui, Chen Sun, Alex Shepard,
  Hartwig Adam, Pietro Perona, and Serge Belongie.
\newblock The inaturalist species classification and detection dataset.
\newblock In {\em Proceedings of the IEEE conference on computer vision and
  pattern recognition}, pages 8769--8778, 2018.

\bibitem{wah2011cub}
C. Wah, S. Branson, P. Welinder, P. Perona, and S. Belongie.
\newblock {The Caltech-UCSD Birds-200-2011 Dataset}.
\newblock {\em Technical Report CNS-TR-2011-001, California Institute of
  Technology}, 2011.

\bibitem{wang2015instre}
S. Wang and S. Jiang.
\newblock {INSTRE: A New Benchmark for Instance-Level Object Retrieval and
  Recognition}.
\newblock {\em ACM Trans. Multimedia Comput. Commun. Appl.}, 2015.

\bibitem{weyand2020google}
T. Weyand, A. Araujo, B. Cao, and J. Sim.
\newblock {Google Landmarks Dataset v2 - A Large-Scale Benchmark for
  Instance-Level Recognition and Retrieval}.
\newblock In {\em Proc. CVPR}, 2020.

\bibitem{wortsman22a}
Mitchell Wortsman, Gabriel Ilharco, Samir~Ya Gadre, Rebecca Roelofs, Raphael
  Gontijo-Lopes, Ari~S Morcos, Hongseok Namkoong, Ali Farhadi, Yair Carmon,
  Simon Kornblith, and Ludwig Schmidt.
\newblock {Model soups: averaging weights of multiple fine-tuned models
  improves accuracy without increasing inference time}.
\newblock In {\em Proc. ICML}, 2022.

\bibitem{wu2022}
C. Wu and S. Maji.
\newblock {How well does CLIP understand texture?}
\newblock In {\em Proc. ECCV Workshops}, 2022.

\bibitem{yang2021dolg}
M. Yang, D. He, M. Fan, B. Shi, X. Xue, F. Li, E. Ding, and J. Huang.
\newblock {DOLG: Single-Stage Image Retrieval with Deep Orthogonal Fusion of
  Local and Global Features}.
\newblock In {\em Proc. ICCV}, 2021.

\bibitem{ypsilantis2021met}
Nikolaos-Antonios Ypsilantis, Noa Garcia, Guangxing Han, Sarah Ibrahimi, Nanne
  Van~Noord, and Giorgos Tolias.
\newblock The met dataset: Instance-level recognition for artworks.
\newblock In {\em Thirty-fifth Conference on Neural Information Processing
  Systems Datasets and Benchmarks Track (Round 2)}, 2021.

\bibitem{zhai2018classification}
Andrew Zhai and Hao-Yu Wu.
\newblock Classification is a strong baseline for deep metric learning.
\newblock {\em arXiv preprint arXiv:1811.12649}, 2018.

\bibitem{zhai2019learning}
A. Zhai, H.-Y. Wu, E. Tzeng, D.~H. Park, and C. Rosenberg.
\newblock {Learning a Unified Embedding for Visual Search at Pinterest}.
\newblock {\em Proc. SIGKDD}, 2019.

\bibitem{zhang2022towards}
M. Zhang, G. Song, Y. Liu, and H. Li.
\newblock {Towards Robust Face Recognition with Comprehensive Search}.
\newblock In {\em Proc. ECCV}, 2022.

\bibitem{zhang2022adaptive}
P. Zhang, H. Dou1, Y. Yu, and X. Li.
\newblock {Adaptive Cross-Domain Learning for Generalizable Person
  Re-Identification}.
\newblock In {\em Proc. ECCV}, 2022.

\bibitem{zhang2022omnibenchmark}
Y. Zhang, Z. Yin, J. Shao, and Z. Liu.
\newblock {Benchmarking Omni-Vision Representation through the Lens of Visual
  Realms}.
\newblock In {\em Proc. ECCV}, 2022.

\bibitem{zheng2015scalable}
L. Zheng, L. Shen, L. Tian, S. Wang, J. Wang, and Q. Tian.
\newblock {Scalable Person Re-identification: A Benchmark}.
\newblock In {\em Proc. ICCV}, 2015.

\bibitem{zhu2022pass}
K. Zhu, H. Guo, T. Yan, Y. Zhu, J. Wang, and M. Tang.
\newblock {PASS: Part-Aware Self-Supervised Pre-Training for Person
  Re-Identification}.
\newblock In {\em Proc. ECCV}, 2022.

\end{thebibliography}
}

\clearpage

\appendix
\section{Appendix}
\subsection{mAP results}
As discussed in Section 3.2, we additionally present results using the mean Average Precision (mAP) metric.
We compute mAP@100, where only the top $100$ retrieved images contribute to the score.
As in~\cite{weyand2020google}, the metric is defined as:
\begin{equation}
    \mathrm{mAP@100} =  \frac{1}{Q} \sum_{q=1}^{Q} \mathrm{AP@100(q)} ,
\end{equation}
where
\vspace{-5px}
\begin{equation}
\mathrm{AP@100(q)} = \frac{1}{\mathrm{min}(m_q, 100)} \sum_{k=1}^{\mathrm{min}(n_q, 100)} \mathrm{P}_q(k) \mathrm{rel}_q(k),
\end{equation}
\noindent where $Q$ is the total number of query images, $m_q$ is the number of index images containing an object in common with the query image $q$ (images from the same class in the index), $n_q$ is the number of predictions made by the system for query $q$ (for our case it is always 100 as we always retrieve 100 images for this metric), $\mathrm{P}_q(k)$ is the precision at rank $k$ for the $q$-th query; and $\mathrm{rel}_q(k)$ is a binary indicator function denoting the relevance of prediction $k$ for the $q$-th query.
Results are presented in Table~\ref{tab:map}.
We observe high correlation between the mAP and the metrics reported in the main paper.
For example, the highest performing method in all cases is obtained with CLIP pre-training and the Oracle Specialist.
The four universal models based on CLIP or IN pre-training perform very similarly: 
their relative ranking remains the same as the one of the mMP@5 metric.
Additionally, for most domains, mMP@5 and mAP agree on the best model.
We conclude that all metrics capture similar trends, while specifically mMP@5 and mAP are very correlated.
To improve metric interpretability and simplicity, as discussed in Section 3.2, we thus decide to establish the two main metrics in our benchmark as mMP@5 and R@1.

\begin{table*}[h]
  \centering
  \scalebox{0.69}{
\begin{tabular}{c|c|c|c|c|c|c|c|c|c}
\hline
& \multicolumn{1}{c|}{Food2k} & \multicolumn{1}{c|}{CARS196}& \multicolumn{1}{c|}{SOP}&\multicolumn{1}{c|}{InShop}&\multicolumn{1}{c|}{iNat}& \multicolumn{1}{c|}{Met}& \multicolumn{1}{c|}{GLDv2}& \multicolumn{1}{c|}{Rp2k}&  \multicolumn{1}{c}{Mean}\\ 
\hline 
Model & \multicolumn{9}{c}{mAP@100} \\
\hline
\multicolumn{10}{c}{\textbf{Off-the-shelf}}\\
\hline
IN (\textbf{\red{768-D}})
& 23.6
& 9.6
& 38.5
& 34.5
& 43.1
& 22.6
& 7.8
& 41.9
& 27.7
\\
CLIP (\textbf{\red{768-D}})
& 21.4
& 29.0
& 39.3
& 36.0
& 24.3
& 23.7 
& 11.6
& 28.6
& 26.7
\\
\hdashline
IN + PCAw
& 13.6
& 5.9
& 26.3
& 18.2
& 27.7
& 9.0
& 3.6
& 28.4
& 16.6
\\
CLIP + PCAw
& 17.1
& 20.0
& 32.2
& 25.9
& 17.9
& 14.0
& 6.4
& 23.4
& 19.6
\\
\hline
\multicolumn{10}{c}{\textbf{Specialists}}\\
\hline
IN+Oracle
& 42.7
& 19.9
& 56.6
& 64.8
& \textbf{49.4}
& 24.1
& 19.4
& 64.8
& 42.7
\\
CLIP+Oracle
& 43.7
& \textbf{40.5}
& 62.6
& 66.2
& 43.9
& \textbf{27.4}
& \textbf{23.1}
& 59.5
& \textbf{45.9}
\\
\hline
\multicolumn{10}{c}{\textbf{Universal models}}\\
\hline
IN+UJCDS
& 44.3
& 15.4
& 51.7
& 58.8
& 48.0
& 4.7
& 17.2
& 65.7
& 38.2
\\
CLIP+UJCDS  
& \textbf{\blue{45.2}}
& 33.4
& 55.6
& 63.1
& 41.3
& 2.6
& \blue{21.2}
& 62.2
& 40.6
\\
\hdashline
IN+UJCRR 
& 42.7
& 23.2
& 61.5
& 73.6
& 48.1
& 5.9
& 12.0
& \textbf{\blue{66.1}}
& 41.6
\\
CLIP+UJCRR
& 43.9
& \blue{39.5}
& \textbf{\blue{65.8}}
& \textbf{\blue{76.7}}
& 40.4
& 5.9
& 15.2
& 61.5
& \blue{43.6}
\\
\hdashline
IN+USCRR 
& 42.2
& 16.7
& 58.2
& 69.5
& \blue{48.6}
& 8.0
& 13.2
& 65.0
& 40.2
\\
CLIP+USCRR
& 41.7
& 36.2
& 61.6
& 71.7
& 40.5
& 9.7
& 15.4
& 62.8
& 42.4
\\
\hdashline
IN+USCSS
& 40.8
& 13.3
& 57.2
& 65.6
& 47.0
& \blue{11.5}
& 16.6
& 64.2
& 39.5
\\ 
CLIP+USCSS
& 42.1
& 33.5
& 62.9
& 70.2
& 42.5
& 8.5
& 20.3
& 61.9
& 42.7
\\
\hline
\end{tabular}
}
  \vspace{-5pt}
  \caption{
  Corresponding mAP@100 for baselines presented in Table 4 of the main paper.
  Color coding follows Table 4.
  \label{tab:map}
  \vspace{-10pt}
  }
\end{table*}

\subsection{Architecture size study}
We study the effect of the ViT architecture size by comparing the performance of ViT-Small, ViT-Base (used in the main paper) and ViT-Large on our evaluation benchmark.
Each of them has larger number of parameters than the previous one, being more memory and computationally expensive.
We compare them by training with the UJCRR method (explained in the main paper), starting from IN pretraining.
Results shown in Table~\ref{tab:ujcrr_architect} justify our choice of ViT-Base as our main backbone; it is a good tradeoff for size and performance, performing as well as the larger ViT-Large, but a lot better than the smaller ViT-Small.
\begin{table*}[h]
  \centering  \scalebox{0.9}{
\begin{tabular}{ccc}
\hline
& \multicolumn{2}{c}{Mean}\\ 
\hline
Model & mMP@5 & R@1\\
\hline
ViT-S (IN) & 48.3 & 58.9 \\
ViT-B (IN) & 52.4 & 62.6 \\
ViT-L (IN) & 52.4 & 62.7 \\
\hline
\end{tabular}
}
  \vspace{-5pt}
  \caption{
  Study for the model architecture.
  All models are finetuned with the UJCRR method described in the main paper.
  \label{tab:ujcrr_architect}
  \vspace{-10pt}
  }
\end{table*}

\subsection{Separate index evaluation}

We perform an evaluation where each domain's queries are tested against the index of the same domain instead of the merged index set, which is the main evaluation of our proposed benchmark.
It corresponds to the setting where an Oracle that restricts the index to images from the same domain as the one of the query image is used.
For this evaluation, only the CLIP pretraining is used.
Results are shown in Table~\ref{tab:separate_db}, and each entry in the table can only be equal or greater than the corresponding one in the main paper.
This is because all cross-domain mistakes are avoided in this setting.
We observe that the universal models and the oracle specialist performs slightly better on average in this setting, with the highest increase being in the Met domain.
This could be caused by the fact that the Met domain contains artworks that can also be considered roughly parts of the other domains as well, \eg clothing pieces, depictions of animals or landmarks in paintings, therefore making it easier to have cross-domain mistakes for Met queries.
Additionally, CLIP+PCAw performance is also a lot higher, showing that naive unsupervised projection with PCA-Whitening produces a lot of cross-domain mistakes.
\begin{table*}[h]
  \centering  \definecolor{LightCyan}{rgb}{0.88,1,1}
\definecolor{LightRed}{rgb}{1.0,0.9,0.9}
\definecolor{LightYellow}{rgb}{1.0,1.0,0.7}
\scalebox{0.53}{
\begin{tabular}{c|cc|cc|cc|cc|cc|cc|cc|cc|cc}
\hline
& \multicolumn{2}{c|}{Food2k} & \multicolumn{2}{c|}{CARS196}& \multicolumn{2}{c|}{SOP}&\multicolumn{2}{c|}{InShop}&\multicolumn{2}{c|}{iNat}& \multicolumn{2}{c|}{Met}& \multicolumn{2}{c|}{GLDv2}& \multicolumn{2}{c|}{Rp2k}&  \multicolumn{2}{c}{Mean}\\ 
\hline
Model & mMP@5 & R@1 & mMP@5 & R@1 & mMP@5 & R@1 & mMP@5 & R@1 & mMP@5 & R@1 & mMP@5 & R@1 & mMP@5 & R@1 & mMP@5 & R@1 & mMP@5 & R@1\\
\hline
\multicolumn{19}{c}{\textbf{Off-the-shelf}}\\
\hline
CLIP (\textbf{\red{768-D}})
& 29.4
& 42.9
& 74.8
& 82.2
& 44.4
& 65.5
& 37.2
& 56.0
& 53.4
& 62.8
& 27.7
& 37.5
& 20.4
& 31.0
& 38.6
& 59.9
& 40.7
& 54.7
\\
\hdashline
CLIP + PCAw
& 29.9
& 41.5
& 67.9
& 76.3
& 40.7
& 61.3
& 39.8
& 57.8
& 52.0
& 60.3
& 19.5
& 25.7
& 16.5
& 23.2
& 40.5
& 59.6
& 38.4
& 50.7
\\
\hline
\multicolumn{19}{c}{\textbf{Specialists}}\\
\hline
CLIP+Oracle
&52.9&64.4&83.3&88.6&67.5&82.1&69.2&86.9&69.3&74.7&33.1&39.8&36.0&47.7&71.1&85.1&60.3&71.2
\\
\hline
\multicolumn{19}{c}{\textbf{Universal models}}\\
\hline
CLIP+UJCDS 
&51.3&62.9&76.1&82.4&58.9&75.7&62.7&80.3&65.0&70.7&5.7&7.0&33.3&45.6&70.2&84.2&52.9&63.6
\\
\hdashline
CLIP+UJCRR
&50.0&62.0&80.3&86.0&68.6&82.7&77.2&90.9&64.6&70.3&9.8&12.3&25.5&36.0&69.8&83.9&55.7&65.5
\\
\hdashline
CLIP+USCRR 
&50.0&61.8&80.4&85.8&66.7&81.7&73.6&89.7&65.7&71.6&12.3&15.9&25.6&36.3&71.9&85.5&55.8&66.0
\\
\hdashline
CLIP+USCSS
&50.1&61.9&78.6&85.0&68.1&82.5&72.5&89.3&67.3&73.2&10.6&14.2&32.6&43.9&71.3&85.1&56.4&66.9
\\
\hline
\end{tabular}
}
  \vspace{-5pt}
  \caption{
  Corresponding separate index evaluation for baselines presented in Table 4 of the main paper.
  \label{tab:separate_db}
  }
\end{table*}

\subsection{Specialists as universal embedding models}
We present evaluation results for specialist models used as universal embeddings in~Table~\ref{tab:specialists}; the highest values for each column are highlighted in bold, the lowest in red, only the CLIP pretraining is used.
As expected, for each domain, the best performing specialist model is the one trained on the corresponding training set, and the best pretraining for that domain corresponds to the one reported in Table 4 of the main paper.
We also note that the best performing models are the specialist models finetuned on Met and Rp2k domains, though the performance of these models is still a lot worse than the best universal model reported in the main paper.
Interestingly, finetuning on the GLDv2 domain performs the worse on average, for both types of pretraining.
\begin{table*}[h]
  \centering
  \definecolor{LightCyan}{rgb}{0.88,1,1}
\definecolor{LightRed}{rgb}{1.0,0.9,0.9}
\definecolor{LightYellow}{rgb}{1.0,1.0,0.7}
\scalebox{0.575}{
\begin{tabular}{c|cc|cc|cc|cc|cc|cc|cc|cc|cc}
\hline
& \multicolumn{2}{|c}{Food2k}& \multicolumn{2}{|c}{CARS196}& \multicolumn{2}{|c}{SOP}&\multicolumn{2}{|c}{InShop}&\multicolumn{2}{|c}{iNat}& \multicolumn{2}{|c}{Met}& \multicolumn{2}{|c}{GLDv2}& \multicolumn{2}{|c}{Rp2k}&  \multicolumn{2}{|c}{Mean}\\ 
\hline \hline
 Model & mMP@5 & R@1 & mMP@5 & R@1 & mMP@5 & R@1 & mMP@5 & R@1 & mMP@5 & R@1 & mMP@5 & R@1 & mMP@5 & R@1 & mMP@5 & R@1 & mMP@5 & R@1\\
\hline
\multicolumn{19}{c}{\textbf{Specialists}}\\
\hline
IN+Food2k
&50.3&63.0&30.9&40.5&29.4&49.1&21.8&34.7&53.5&60.2&7.8&7.0&8.8&13.6&40.5&61.9&30.1&41.2
\\
CLIP+Food2k
&
\textbf{51.3}
&
\textbf{63.5}
&
70.1
&
78.8
&
29.6
&
49.0
&
25.5
&
40.9
&
42.9
&
50.3
&
4.6
&
6.0
&
16.7
&
24.4
&
34.9
&
56.0
&
34.4
&
46.1
\\
\hdashline
IN+CARS196
&19.6&29.4&62.4&72.0&\alert{26.1}&\alert{45.1}&\alert{20.3}&\alert{33.1}&54.8&61.3&11.6&16.6&8.1&12.7&38.6&60.1&30.2&41.3
\\
CLIP+CARS196
&19.0&28.7&\textbf{82.6}&\textbf{88.4}&29.1&48.4&24.5&39.9&42.5&50.1&10.0&13.8&14.7&22.2&27.6&46.5&31.2&42.2
\\
\hdashline
IN+SOP 
&13.1&20.9&22.2&32.0&61.2&78.3&29.3&45.9&44.9&52.6&\alert{2.5}&\alert{3.1}&\alert{5.6}&\alert{9.1}&44.0&66.1&27.8&38.5
\\
CLIP+SOP 
&10.7&17.9&44.3&56.5&\textbf{66.2}&\textbf{81.4}&32.1&50.0&30.2&38.3&3.0&4.3&8.6&13.5&37.0&59.2&29.0&40.1
\\
\hdashline
IN+InShop 
&13.4&21.6&23.6&32.8&33.5&54.7&66.6&86.1&45.8&53.2&5.5&7.2&6.8&11.8&40.0&62.2&29.4&41.2
\\
CLIP+InShop 
&13.1&21.0&61.0&70.6&31.7&51.8&\textbf{67.8}&\textbf{86.2}&35.0&42.5&6.3&8.3&12.1&19.8&31.0&52.1&32.2&44.0
\\
\hdashline
IN+iNat
&24.1&34.8&34.4&44.0&29.6&49.3&24.6&39.0&\textbf{70.0}&\textbf{75.1}&13.8&20.7&10.2&16.2&41.8&62.6&31.1&42.7
\\
CLIP+iNat
&17.6&27.5&61.4&71.0&30.6&50.1&27.4&43.4&67.1&72.7&10.1&13.6&11.3&16.9&34.0&54.7&32.4&43.7
\\
\hdashline
IN+Met 
&14.7&23.7&28.5&39.4&38.0&59.5&33.8&52.8&43.2&51.0&21.7&25.9&9.6&16.1&48.6&70.3&29.8&42.3
\\
CLIP+Met
&16.1&25.4&59.6&70.2&43.8&64.5&40.5&61.5&36.9&45.1&\textbf{25.7}&\textbf{30.8}&16.1&24.6&44.6&66.9&35.4&\textbf{48.6}
\\
\hdashline
IN+GLDv2 
&12.7&19.9&\alert{13.7}&\alert{22.6}&36.6&57.8&25.7&40.1&43.5&50.9&3.2&4.3&31.6&43.8&41.2&63.3&26.0&37.8
\\
CLIP+GLDv2
&\alert{9.7}&\alert{16.4}&23.4&33.3&33.3&53.6&22.3&36.5&\alert{26.1}&\alert{33.7}&3.3&4.4&\textbf{35.6}&\textbf{46.7}&\alert{26.8}&\alert{46.3}&\alert{22.6}&\alert{33.9}
\\
\hdashline
IN+Rp2k
&21.4&31.8&34.8&45.7&34.7&56.3&27.1&42.8&54.4&61.6&14.3&19.8&10.4&17.4&\textbf{73.6}&\textbf{87.2}&33.8&45.3
\\
CLIP+Rp2k
&19.0&29.1&62.8&71.6&34.6&55.6&29.3&46.0&38.9&47.1&15.5&20.5&15.4&25.2&69.6&84.6&\textbf{35.6}&47.5
\\
\hline
\end{tabular}
}
  \vspace{-5pt}
  \caption{
  Results for specialist models when used as universal embeddings on our benchmark.
  Model column has the format : \{Pretraining\}+\{Finetuning dataset\}.
  \label{tab:specialists}
  \vspace{-10pt}
  }
\end{table*}

\subsection{PCA-Whitening vs Random Projection}
We present a comparison of PCA-whitening as a means to reduce the dimensionality of the off-the-shelf embeddings shown in the Table 4 of the main paper versus Random Linear projection to 64-D, in Table~\ref{tab:pcaw_vs_rand_proj}.
Results on the original dimensionality results are also shown for reference.
The random linear projection results are averaged over 3 seeds.
PCA-Whitening has been trained on the union of subsets of $\sim$9k images of each domain.
We observe that for ImageNet pretraining, the random projection performs better on average than PCA-Whitening, while for CLIP pretraining it underperforms the former.
\begin{table*}[h]
  \centering
  \definecolor{LightCyan}{rgb}{0.88,1,1}
\definecolor{LightRed}{rgb}{1.0,0.9,0.9}
\definecolor{LightYellow}{rgb}{1.0,1.0,0.7}
\scalebox{0.526}{
\begin{tabular}{c|cc|cc|cc|cc|cc|cc|cc|cc|cc}
\hline
& \multicolumn{2}{c|}{Food2k} & \multicolumn{2}{c|}{CARS196}& \multicolumn{2}{c|}{SOP}&\multicolumn{2}{c|}{InShop}&\multicolumn{2}{c|}{iNat}& \multicolumn{2}{c|}{Met}& \multicolumn{2}{c|}{GLDv2}& \multicolumn{2}{c|}{Rp2k}&  \multicolumn{2}{c}{Mean}\\ 
\hline
Model & mMP@5 & R@1 & mMP@5 & R@1 & mMP@5 & R@1 & mMP@5 & R@1 & mMP@5 & R@1 & mMP@5 & R@1 & mMP@5 & R@1 & mMP@5 & R@1 & mMP@5 & R@1\\
\hline
\multicolumn{19}{c}{\textbf{Off-the-shelf}}\\
\hline
IN (\textbf{\red{768-D}})
& 31.1
& 44.1
& 41.4
& 54.1
& 43.7
& 65.6
& 35.5
& 53.9
& 67.1
& 74.2
& 21.1
& 30.8
& 14.8
& 25.2
& 52.9
& 74.3
& 38.4
& 52.8
\\
CLIP (\textbf{\red{768-D}})
& 29.4
& 42.9
& 74.7
& 82.2
& 44.2
& 65.4
& 37.2
& 56.0
& 52.4
& 61.9
& 21.4
& 28.5
& 20.4
& 31.0
& 38.6
& 59.9
& 39.8
& 53.5
\\
\hdashline
IN+PCAw 
& 19.1
& 29.1
& 29.0
& 37.8
& 30.5
& 51.2
& 19.6
& 31.6
& 50.9
& 57.9
& 8.0
& 11.0
& 8.3
& 13.2
& 37.6
& 57.8
& 25.4
& 36.2
\\
CLIP+PCAw
& 23.4
& 34.6
& 62.8
& 72.7
& 36.5
& 57.0
& 27.0
& 41.8
& 42.7
& 51.1
& 12.1
& 15.8
& 11.9
& 17.6
& 32.0
& 51.8
& 31.0
& 42.8 
\\
\hdashline
IN+Rand.Proj.
& 19.4\scriptsize{\scriptsize{$\pm$0.5}}
& 29.5\scriptsize{$\pm$0.8}
& 31.0\scriptsize{$\pm$0.3}
& 41.7\scriptsize{$\pm$0.5}
& 33.1\scriptsize{$\pm$0.1}
& 54.5\scriptsize{$\pm$0.3}
& 25.7\scriptsize{$\pm$0.6}
& 40.4\scriptsize{$\pm$0.3}
& 54.4\scriptsize{$\pm$0.2}
& 61.5\scriptsize{$\pm$0.2}
& 8.8\scriptsize{$\pm$0.5}
& 12.2\scriptsize{$\pm$0.7}
& 8.7\scriptsize{$\pm$0.2}
& 14.8\scriptsize{$\pm$0.7}
& 38.7\scriptsize{$\pm$0.1}
& 60.0\scriptsize{$\pm$0.2}
& 27.5\scriptsize{$\pm$0.1}
& 39.3\scriptsize{$\pm$0.1}
\\
CLIP+Rand.Proj.
& 18.1\scriptsize{$\pm$0.7}
& 28.5\scriptsize{$\pm$0.8}
& 61.7\scriptsize{$\pm$1.3}
& 71.7\scriptsize{$\pm$0.8}
& 34.5\scriptsize{$\pm$0.3}
& 55.0\scriptsize{$\pm$0.5}
& 26.8\scriptsize{$\pm$0.5}
& 42.0\scriptsize{$\pm$0.8}
& 41.3\scriptsize{$\pm$0.2}
& 49.8\scriptsize{$\pm$0.2}
& 9.7\scriptsize{$\pm$0.7}
& 13.2\scriptsize{$\pm$1.0}
& 12.5\scriptsize{$\pm$0.5}
& 18.5\scriptsize{$\pm$1.2}
& 29.3\scriptsize{$\pm$0.6}
& 47.7\scriptsize{$\pm$0.5}
& 29.2\scriptsize{$\pm$0.5}
& 40.8\scriptsize{$\pm$0.7}
\\

\hline
\end{tabular}
}
  \vspace{-5pt}
  \caption{
  Comparison of PCA-Whitening vs Random linear projection.
  For the latter, the average of 3 seeds is shown.
  \label{tab:pcaw_vs_rand_proj}
  \vspace{-10pt}
  }
\end{table*}

\subsection{Visual presentation of all domains}
\begin{figure*}[h]
\begin{center}
\includegraphics[scale = 0.64]{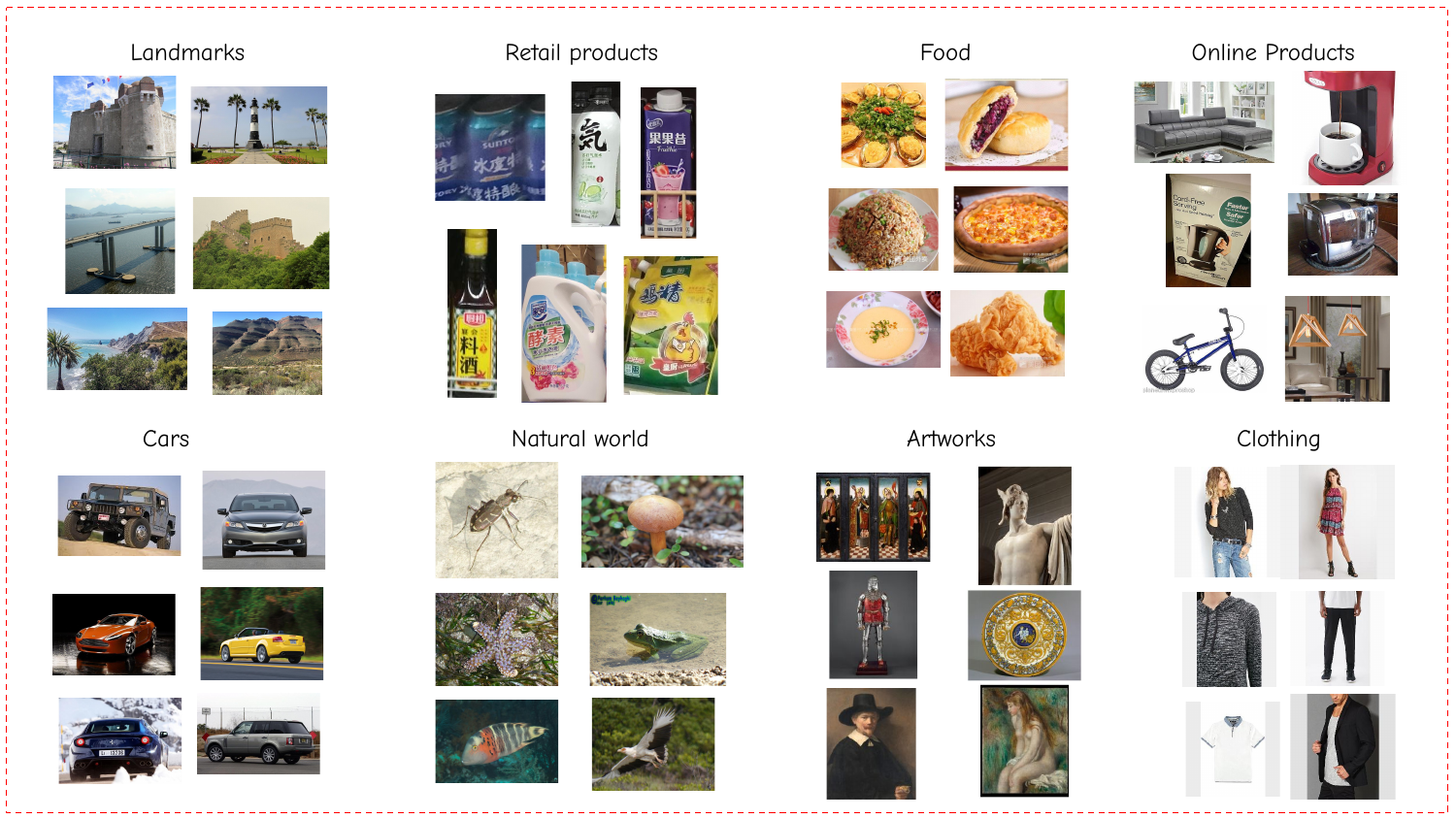}
\caption{
Example images from domains covered by the proposed UnED dataset.
\label{fig:all_domains}}
\end{center}
\end{figure*}

In Figure~\ref{fig:all_domains} we show a collective presentation of example images from the different domains the UnED dataset covers.

\end{document}